\newif\ifarxiv
\arxivtrue   % arxiv
\documentclass[sigconf]{acmart}

\AtBeginDocument{%
  }

\copyrightyear{2026}
\acmYear{2026}
\setcopyright{cc}
\setcctype{by}
\acmConference[MM '26]{Proceedings of the 34th ACM International Conference on Multimedia}{November 10--14, 2026}{Rio de Janeiro, Brazil}
\acmBooktitle{Proceedings of the 34th ACM International Conference on Multimedia (MM '26), November 10--14, 2026, Rio de Janeiro, Brazil}
\acmDOI{10.1145/3767308.3835747}
\acmISBN{979-8-4007-2213-4/2026/11}

\usepackage{subcaption}
\usepackage{multirow}
\usepackage{colortbl}
\usepackage{pifont}
\usepackage{enumitem}

\ifarxiv
  \usepackage{tcolorbox}
\fi

\newcommand{\cmark}{\textcolor{green!60!black}{\ding{51}}}
\newcommand{\xmark}{\textcolor{red!80!black}{\ding{55}}}

\usepackage[table]{xcolor}
\begin{document}

\title{PROVE: A Perceptual RemOVal cohErence Benchmark for Visual Media}


% \author{
%     Fuhao Li\textsuperscript{*},
%     Shaofeng You\textsuperscript{*},
%     Jiagao Hu\textsuperscript{*},
%     Yu Liu,
%     Yuxuan Chen,
%     Zepeng Wang,
%     Fei Wang, \\
%     Daiguo Zhou,
%     Jian Luan
%     }
% \email{{lifuhao5, youshaofeng, hujiagao}@xiaomi.com}
% \affiliation{%
%   \institution{MiLM Plus, Xiaomi Inc.}
%   % \country{China}
% }

\author{Fuhao Li}
\orcid{0009-0003-4830-2171}
\authornote{These authors contributed equally to this work.}
\affiliation{%
  \institution{MiLM Plus, Xiaomi Inc.}
  \city{Wuhan}
  \country{China}
}
\email{lifuhao5@xiaomi.com}

\author{Shaofeng You}
\orcid{0009-0009-0729-7998}
\authornotemark[1]
\ifarxiv
\affiliation{%
  \institution{MiLM Plus, Xiaomi Inc.}
  \country{Zhejiang University}
}
\email{youshaofeng@zju.edu.cn}
\else
\affiliation{%
  \institution{MiLM Plus, Xiaomi Inc.}
  \city{Wuhan}
  \country{China}
}
\email{youshaofeng@xiaomi.com}
\fi

\author{Jiagao Hu}
\orcid{0000-0002-8439-8903}
\authornotemark[1]
\correspondingauthor
\affiliation{%
  \institution{MiLM Plus, Xiaomi Inc.}
  \city{Wuhan}
  \country{China}
}
\email{hujiagao@xiaomi.com}

\author{Yuxuan Chen}
\orcid{0000-0003-1341-1354}
\affiliation{%
  \institution{MiLM Plus, Xiaomi Inc.}
  \city{Wuhan}
  \country{China}
}
\email{chenyuxuan7@xiaomi.com}

\author{Zepeng Wang}
\orcid{0000-0002-2366-8455}
\affiliation{%
  \institution{MiLM Plus, Xiaomi Inc.}
  \city{Wuhan}
  \country{China}
}
\email{wangzepeng5@xiaomi.com}

\author{Yu Liu}
\orcid{0009-0008-0407-1056}
\affiliation{%
  \institution{MiLM Plus, Xiaomi Inc.}
  \city{Wuhan}
  \country{China}
}
\email{liuyu46@xiaomi.com}

\author{Fei Wang}
\orcid{0009-0009-5698-396X}
\affiliation{%
  \institution{MiLM Plus, Xiaomi Inc.}
  \city{Wuhan}
  \country{China}
}
\email{wangfei11@xiaomi.com}

\author{Daiguo Zhou}
\orcid{0009-0009-9351-2067}
\affiliation{%
  \institution{MiLM Plus, Xiaomi Inc.}
  \city{Wuhan}
  \country{China}
}
\email{zhoudaiguo@xiaomi.com}

\author{Jian Luan}
\orcid{0000-0002-2383-226X}
\affiliation{%
  \institution{MiLM Plus, Xiaomi Inc.}
  \city{Beijing}
  \country{China}
}
\email{luanjian@xiaomi.com}

\renewcommand{\shortauthors}{Fuhao Li et al.}

\begin{abstract}

Evaluating object removal in images and videos remains challenging because the task is inherently one-to-many, yet existing metrics frequently disagree with human perception. Full-reference metrics reward copy-paste behaviors over genuine erasure; no-reference metrics suffer from systematic biases such as favoring blurry results; and global temporal metrics are insensitive to localized artifacts within edited regions.
To address these limitations, we propose \textbf{RC} (Removal Coherence), a pair of perception-aligned metrics: RC-S, which measures spatial coherence via sliding-window feature comparison between masked and background regions, and RC-T, which measures temporal consistency via distribution tracking within shared restored regions across adjacent frames. To validate RC and support community benchmarking, we further introduce \textbf{\mbox{PROVE-Bench}}, a two-tier real-world benchmark comprising \mbox{PROVE-M}, an 80-video paired dataset with motion augmentation, and \mbox{PROVE-H}, a 100-video challenging subset without ground truth. Together, RC metrics and PROVE-Bench form the \textbf{PROVE} (Perceptual RemOVal cohErence) evaluation framework for visual media. Experiments across diverse image and video benchmarks demonstrate that RC achieves substantially stronger alignment with human judgments than existing evaluation protocols.
Project page: \url{https://xiaomi-research.github.io/prove/}.

\end{abstract}






\maketitle

\section{Introduction}
\label{sec:intro_2}

Object removal in images and videos aims to erase user-specified objects and seamlessly restore the occluded background in a natural and coherent manner. As a core technology in content editing, scene cleanup, and post-production, this field has seen remarkable progress driven by diffusion models. Yet as generation quality continues to improve, evaluation has emerged as a critical bottleneck: existing metrics and benchmarks often fail to faithfully reflect the true quality of object removal results.

\begin{figure}[t]
  \centering
  \includegraphics[width=\linewidth]{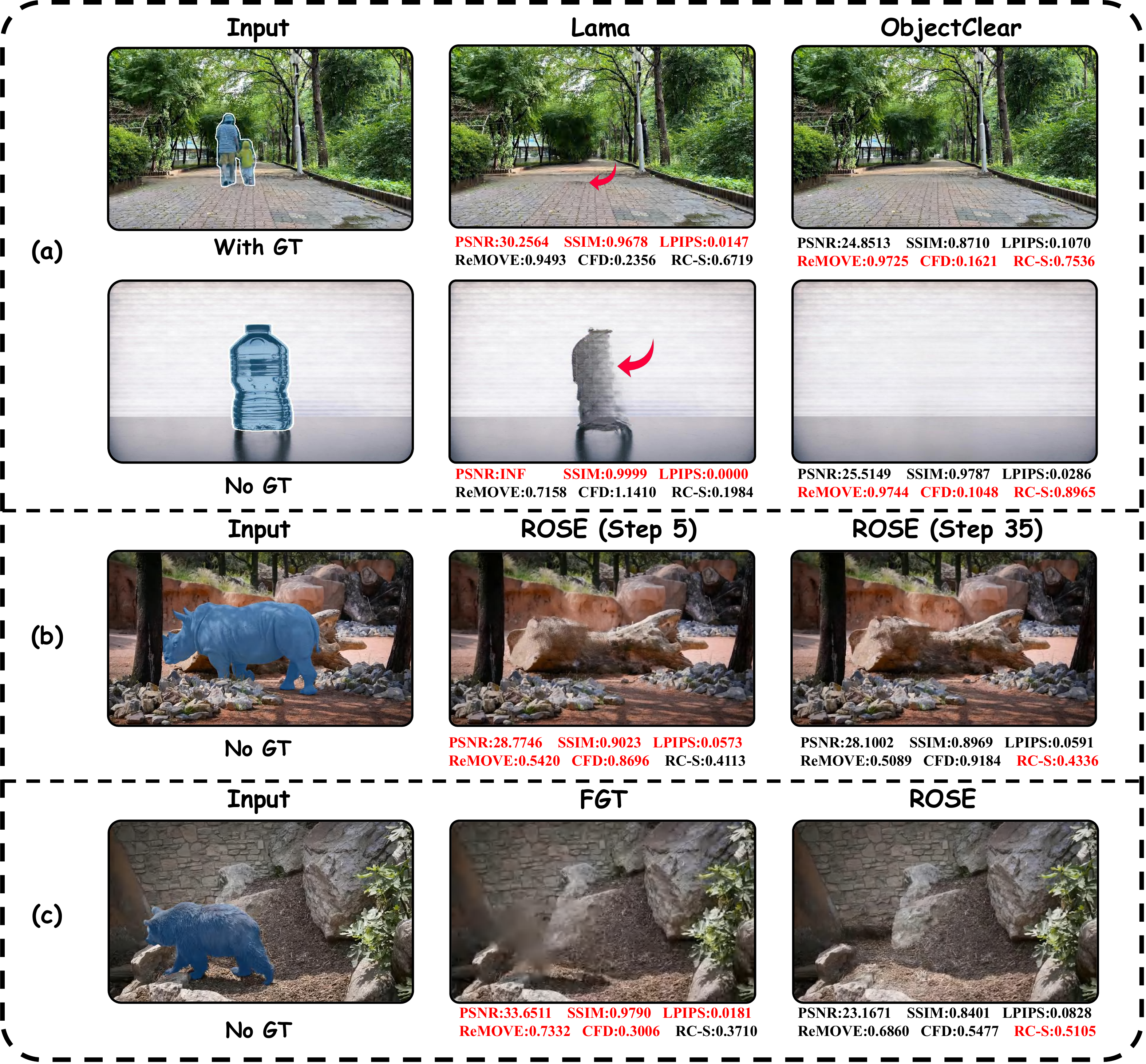}
  \caption{Examples of metric bias in object removal evaluation. (a) LaMa~\cite{suvorov2022resolution} and ObjectClear~\cite{zhao2026objectclear} with and without paired GT. (b) ROSE~\cite{miao2025rose} with different diffusion steps. (c) FGT~\cite{zhang2022flow} and ROSE~\cite{miao2025rose}. FR and NR scores are shown below each result; \textcolor{red}{red} marks metric-preferred results that conflict with visual perception.}
  \label{fig:metric_bias}
\end{figure}

The difficulty of evaluating object removal stems from its inherently \textit{ill-posed one-to-many} nature~\cite{chen2024assessing}: multiple perceptually plausible restorations may exist for the same erased region, meaning that a unique ground truth (GT) cannot be defined. This undermines traditional full-reference (FR) metrics such as PSNR~\cite{hore2010image}, SSIM~\cite{wang2004image}, and LPIPS~\cite{zhang2018unreasonable}, which assume strict point-to-point correspondence to a single reference and consequently reward conservative outputs over perceptually realistic ones, as shown in Fig.~\ref{fig:metric_bias}(a).

No-reference (NR) metrics are more practically attractive as they do not rely on any GT. Representative methods include ReMOVE~\cite{chandrasekar2024remove} and CFD~\cite{yu2025omnipaint}. However, our experiments reveal that these metrics suffer from systematic blind spots: they frequently assign inflated scores to blurry outputs and incorrectly penalize structurally sound restorations in complex occlusion scenarios, as shown in Fig.~\ref{fig:metric_bias}(b)(c).

In the video domain, temporal consistency introduces an additional evaluation dimension that existing metrics handle poorly. Temporal Consistency (TC)~\cite{zhang2024avid} and Temporal Flickering (TF)~\cite{huang2024vbench} are computed over full-frame features and thus dominated by unchanged background regions, failing to detect localized artifacts within the removed regions — precisely where object removal most commonly fails.

Beyond metrics, reliable evaluation also depends on benchmarks, which are particularly difficult to construct for video removal. Publicly available datasets fall into two categories. Synthetic datasets such as Movies~\cite{lin2023omnimatterf} and ROSE-Bench~\cite{miao2025rose} provide precise masks and reference videos but cannot fully reproduce real-world complexity.
Real-world datasets such as DAVIS~\cite{pont20172017} offer greater realism but lack paired target-free videos, making it difficult to establish even a reference baseline for quantitative comparison across models and metrics. No existing benchmark simultaneously offers real-world authenticity and paired reference videos for controlled evaluation.

To bridge these critical gaps, we propose \textbf{PROVE} (\textbf{P}erceptual \textbf{R}em\textbf{OV}al coh\textbf{E}rence), a unified evaluation framework comprising two novel perception-aligned metrics, Removal Coherence Spatial (\textbf{RC-S}) and Removal Coherence Temporal (\textbf{RC-T}), together with a multi-tier benchmark suite, \textbf{PROVE-Bench}.

RC-S evaluates spatial coherence by cropping each target region, extracting deep features, and applying a sliding-window Maximum Mean Discrepancy (MMD)~\cite{gretton2012kernel} to compare feature distributions inside and outside the removed region, enabling fine-grained detection of local spatial incoherence. RC-T extends this design to the temporal domain by jointly cropping adjacent frames under a shared union mask and measuring feature distribution drift exclusively within the intersected restored regions, yielding sensitive detection of local temporal instability.
PROVE-Bench consists of two complementary subsets. PROVE-M provides precisely aligned input–mask–ground-truth video triplets captured in real-world scenes, and PROVE-H complements this with 100 challenging real-world videos without ground truth, targeting extreme scenarios such as crowds, fast motion, and complex reflections to stress-test model generalization ability.

Overall, our main contributions are three-fold:
\begin{itemize}[leftmargin=1em, nosep]
    \item We systematically diagnose the failure modes of existing evaluation metrics for object removal — including the copy-paste bias in FR metrics, the blur-favoring bias in NR metrics, and the regional insensitivity of global temporal metrics — providing a rigorous empirical foundation for rethinking removal evaluation.
    \item We propose \textbf{RC-S} and \textbf{RC-T}, two perception-aligned metrics that quantify local spatial coherence and temporal consistency of object removal results via a unified sliding-window distribution matching framework, achieving substantially stronger alignment with human judgments than existing protocols.
    \item We construct PROVE-Bench, a two-tier benchmark suite that uniquely combines PROVE-M, a real-world paired dataset, and PROVE-H, a challenging GT-free dataset, providing complementary evaluation support for both rigorous quantitative comparison and stress-testing under unconstrained real-world conditions.
\end{itemize}

\section{Related Work}
\label{sec:related_2}

\subsection{Image and Video Object Removal}
Image and video object removal erases target regions and reconstructs occluded backgrounds. Earlier convolutional, adversarial, and motion-propagation methods~\cite{suvorov2022resolution,zhang2022flow,zhou2023propainter,yildirim2023diverse} work in relatively simple settings but struggle with large holes and object-related effects such as shadows and reflections. Recent diffusion-based methods~\cite{li2025diffueraser,miao2025rose,zi2025minimax,jiang2025vace,lee2025generative,zhao2026objectclear,wei2025omnieraser} improve restoration realism and the handling of complex object interactions.

\subsection{Evaluation Metrics for Object Removal}
Existing evaluation methods include full-reference (FR), no-reference (NR), and temporal metrics.

\noindent\textbf{Full-Reference (FR) Metrics:}
FR metrics compare outputs with ground truth (GT) using pixel-level measures such as PSNR~\cite{hore2010image} and SSIM~\cite{wang2004image}, perceptual measures such as LPIPS~\cite{zhang2018unreasonable}, or distributional measures such as FID~\cite{heusel2017gans} and CMMD~\cite{jayasumana2024rethinking}. For videos, they are typically computed frame by frame.

\noindent\textbf{No-Reference (NR) Metrics:}
For settings without GT, ReMOVE~\cite{chandrasekar2024remove} measures feature similarity between the inpainted region and its context, while CFD~\cite{yu2025omnipaint} uses SAM~\cite{kirillov2023segment} to penalize potential hallucinations. TokSim~\cite{kushwaha2026object} extends NR evaluation to video by combining temporal coherence, structural divergence, and spatial blending.

\noindent\textbf{Temporal Metrics:}
Temporal Consistency (TC)~\cite{zhang2024avid} measures cosine similarity between CLIP embeddings of adjacent frames, whereas Temporal Flickering (TF)~\cite{huang2024vbench} measures their pixel-level difference. Despite their widespread use, these metrics can yield counter-intuitive results for object removal, as analyzed in Sec.~\ref{sec:metric_analysis}.

\subsection{Video Object Removal Benchmarks}
Existing benchmarks are either synthetic paired datasets or real-world unpaired datasets. Kubric~\cite{wu2022d} provides paired videos but has relatively simple scenes and lighting; Movies~\cite{lin2023omnimatterf} adds complex lighting and non-rigid motion; and ROSE-Bench~\cite{miao2025rose} targets object-related effects such as shadows and reflections. DAVIS~\cite{pont20172017} provides real-world videos but was designed for segmentation, lacks diverse removal scenarios, and provides no paired target-free ground truth.
\section{Limitations of Existing Metrics}
\label{sec:metric_analysis}

\begin{figure}
  \centering
  \includegraphics[width=\linewidth]{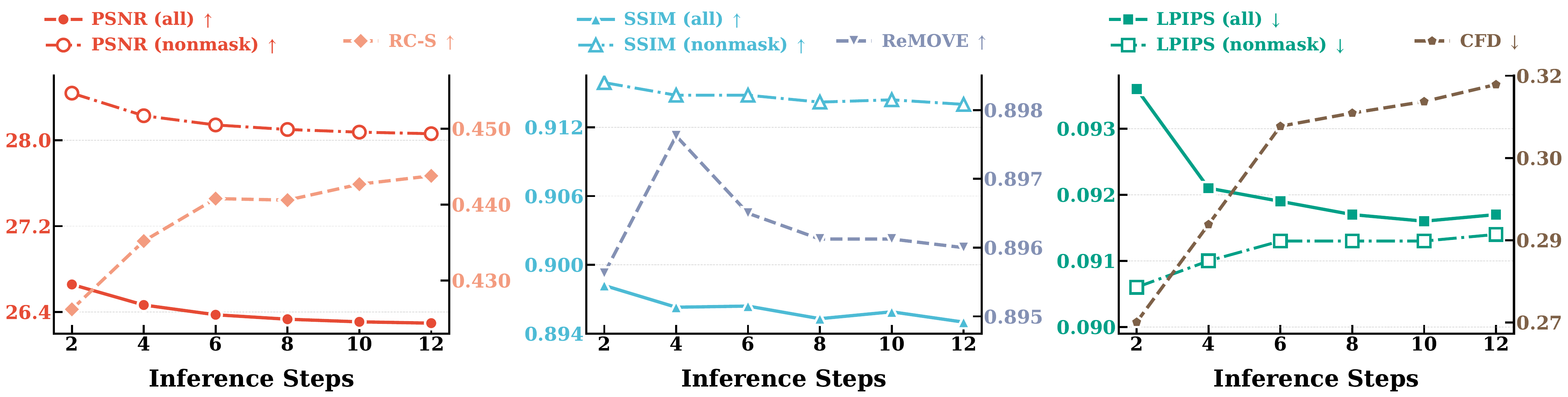}
  \caption{Metric responses to different inference steps for Minimax-Remover~\cite{zi2025minimax} on ROSE-Bench.}
  \label{fig:1}
\end{figure}

\begin{figure}
  \centering
  \includegraphics[width=\linewidth]{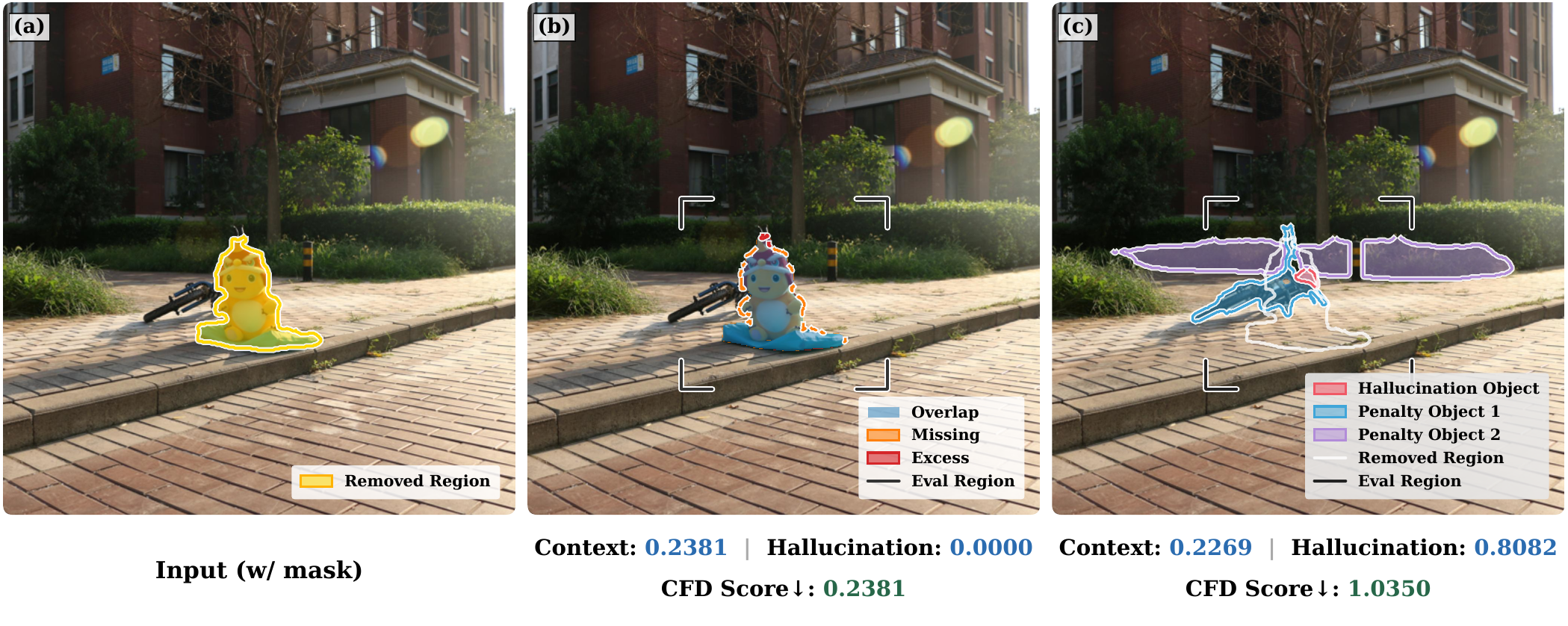}
  \caption{Failure example of CFD. (a) Input image with mask. (b) CFD visualization on the input image, where the residual foreground is not penalized. (c) CFD visualization on the GT, where the restored bicycle seat is misclassified as a hallucination, yielding a worse score than the unedited input.}
  \label{fig:3}
\end{figure}

\subsection{Full-Reference Metrics}
\label{sec:fr_limitations}
FR metrics (e.g. PSNR~\cite{hore2010image},
SSIM~\cite{wang2004image}, LPIPS~\cite{zhang2018unreasonable}) fundamentally struggle with the ill-posed nature of object removal. Through empirical evaluation, we identify two counter-intuitive failure modes that further expose their unreliability in the object removal tasks.

\textbf{``Copy-Paste'' Bias.} FR metrics are fundamentally biased toward ``copy-paste'' behaviors rather than genuine erasure, often rewarding non-diffusion models that mechanically preserve the background. This leads to two paradoxes as shown in Fig.~\ref{fig:metric_bias}(a). When GT is available, residual effects such as shadows occupy a negligible pixel area and thus incur minimal penalty. When GT is unavailable, FR metrics are usually computed solely on unmasked regions~\cite{zi2025minimax, kushwaha2026object}; the flaw is further exacerbated: even significant removal failures may yield near-perfect scores, producing evaluations that clearly contradict human perception~\cite{sun2023privacy, ghildyal2023attacking}.

\textbf{``Regression to the Mean'' Bias.} Counter-intuitively, reducing the number of diffusion inference steps often \emph{improves} FR scores despite severe degradation in visual quality (Fig.~\ref{fig:metric_bias}(b) and Fig.~\ref{fig:1}). For pixel-wise metrics such as PSNR and SSIM, this stems from the mathematical ``regression to the mean'' effect~\cite{barnett2005regression}: by heavily penalizing pixel variance, they inherently favor smoothed approximations over realistic high-frequency details~\cite{blau2018perception, sajjadi2017enhancenet, whang2022deblurring,wang2025traversing}.
Although LPIPS adopts deep features to better reflect human perception, its underlying networks exhibit poor shift-equivariance, and its distance metric enforces strict point-to-point feature matching, making it highly sensitive to imperceptible spatial shifts. Moreover, its patch-based computation further imposes a limited receptive field~\cite{ghildyal2022shift}, preventing it from capturing global semantic coherence~\cite{sun2023privacy}. These phenomena are consistent with the well-known perception-distortion tradeoff~\cite{blau2018perception}.

Together, these failure modes reveal that FR metrics are fundamentally mismatched with the objectives of object removal: they overly reward reference similarity while severely under-penalizing task-critical local failures.

\begin{figure}
  \centering
  \includegraphics[width=\linewidth]{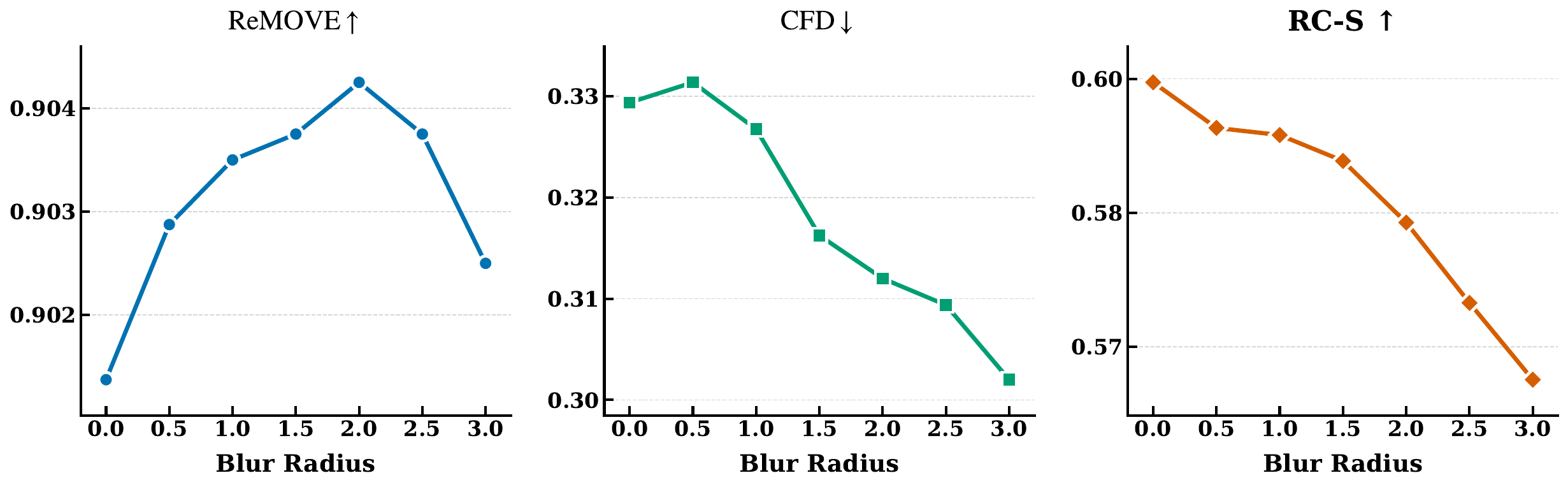}
  \caption{Sensitivity of ReMOVE, CFD, and the proposed RC-S to increasing Gaussian blur in the masked region.}
  \label{fig:blur_sensitivity}
\end{figure}

\begin{figure}
    \centering
    \includegraphics[width=\linewidth]{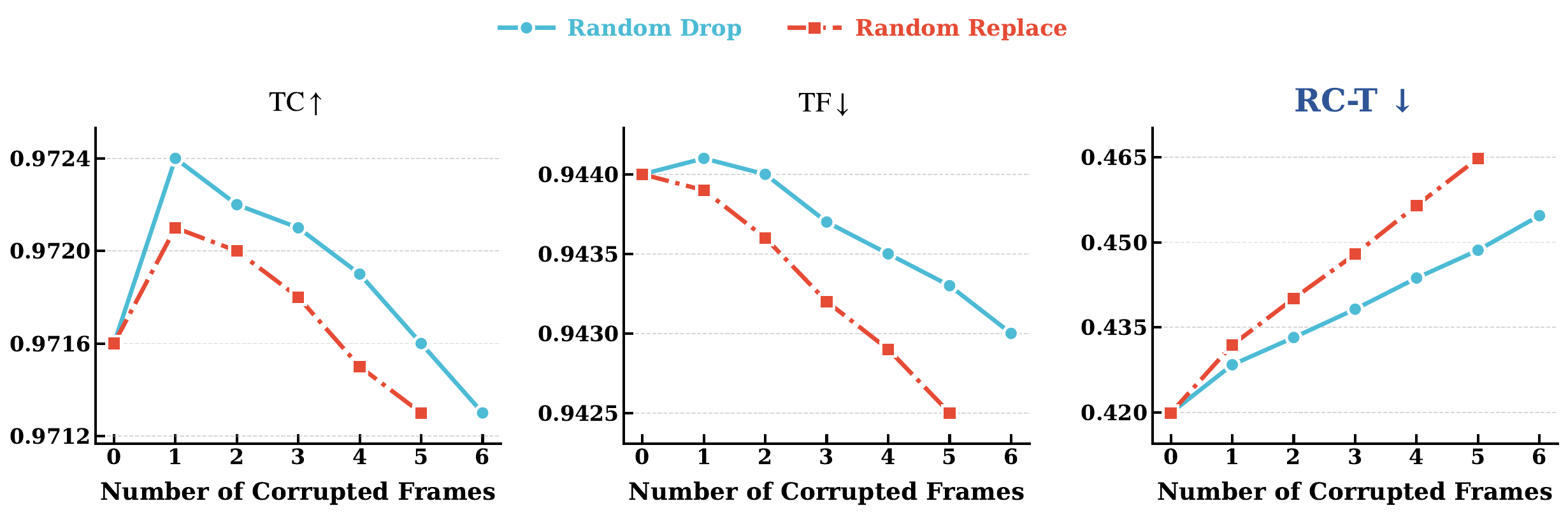}
    \caption{\textbf{Sensitivity of temporal metrics to Random Drop and Random Replace corruptions on DAVIS.}}
    \label{fig:both_datasets}
\end{figure}

\subsection{No-Reference Metrics}
\label{sec:nr_limitations}

NR metrics are more attractive for object removal as they require no GT reference.
However, representative methods such as ReMOVE~\cite{chandrasekar2024remove} and
CFD~\cite{yu2025omnipaint} still suffer from systematic biases that undermine their
reliability.

\textbf{``Blur is Clean'' Bias.}
As shown in Fig.~\ref{fig:metric_bias}(c) and Fig.~\ref{fig:1}, both ReMOVE and CFD consistently assign higher scores to blurrier results — whether from reducing diffusion steps or comparing earlier methods (e.g., FGT~\cite{zhang2022flow}) against more advanced ones (e.g., ROSE~\cite{miao2025rose}). To further verify this, we conduct a controlled experiment on ROSE-Bench by progressively applying Gaussian blur within the masked region. As shown in Fig.~\ref{fig:blur_sensitivity}, neither metric decreases with increasing blur; both eventually surpass their unblurred baselines. This is not an isolated anomaly but a structural consequence of measuring feature similarity via global or first-order statistics (e.g., cosine similarity) without localized comparison. Any metric relying on such aggregation strategies inherits this susceptibility, as we analyze further in Sec.~\ref{sec:ablation}.

\begin{figure*}
  \centering
  \includegraphics[width=1.0\linewidth]{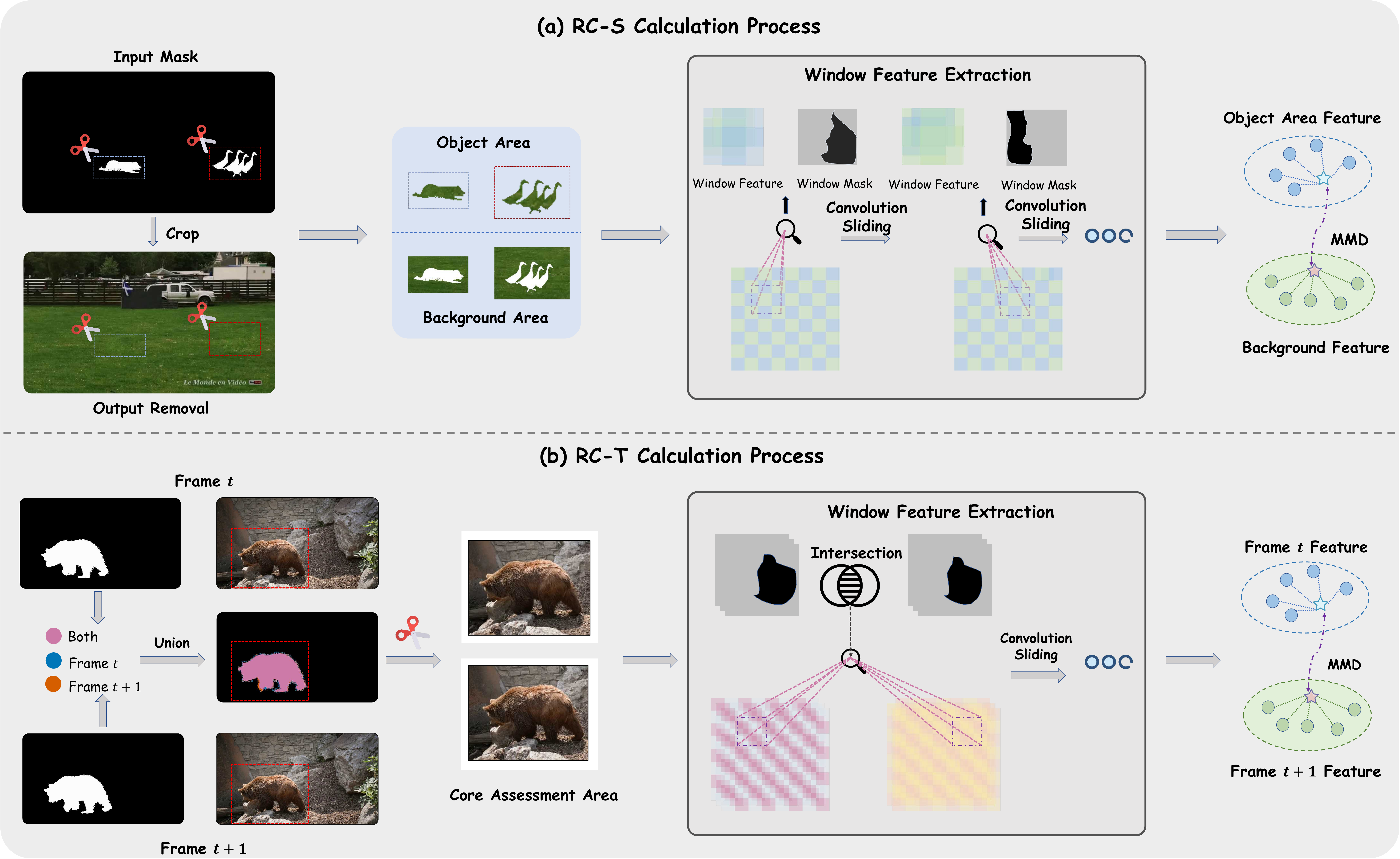}
  \caption{Overview of the proposed RC metrics. (a) RC-S measures intra-frame spatial coherence by comparing masked and background feature distributions within sliding windows. (b) RC-T measures inter-frame temporal consistency by comparing restored-region feature distributions across adjacent frames under union-based cropping and intersection-based evaluation.}
  \label{fig:method}
\end{figure*}

\textbf{``Original is Better'' Bias.} CFD decomposes its score into \emph{Context Coherence} and \emph{Hallucination Penalty} to address ReMOVE's insensitivity to hallucinated objects. However, we find that both components can incorrectly favor the unedited input over a perfectly removed result.
As shown in Fig.~\ref{fig:3}, CFD uses SAM to detect isolated ``nested'' masks as hallucinations — a heuristic that is highly vulnerable in occlusion scenarios. When a foreground object (e.g., a doll) is removed to reveal an occluded background structure (e.g., a bicycle), naturally restored elements (e.g., the bicycle seat) are frequently misclassified as hallucinations. Conversely, unremoved targets often evade penalization by being miscategorized as ``boundary extensions'' due to rigid boundary thresholds. Additional analyses of existing NR metrics are provided in the appendix
\ifarxiv
  (Sec.~\ref{sec:CFD}).
\else
  (Sec.~A.3).
\fi

\subsection{Temporal Metrics}
\label{sec:lim_tm}

Temporal consistency is a core quality criterion in video object removal. Widely adopted metrics such as Temporal Consistency (TC)~\cite{zhang2024avid} and Temporal Flickering (TF)~\cite{huang2024vbench} assess inter-frame stability by operating on global frame-level features. However, since the removed region typically occupies only a small fraction of the frame, this global aggregation dilutes or entirely masks temporal errors localized within the edited area.

To expose this limitation, we conduct a sensitivity analysis. Starting from high-quality results as baselines, we introduce two types of synthetic temporal corruption: \textbf{Random Drop}, which removes intermediate frames to simulate motion jumps and temporal incoherence, and \textbf{Random Replace}, which substitutes frames with temporally distant ones to simulate severe flickering or abrupt content changes. We then progressively increase the number of corrupted frames and measure each metric.

Figure ~\ref{fig:both_datasets} shows the results on DAVIS dataset, TC and TF remain highly insensitive to both types of corruption — their scores not only fail to reflect the introduced artifacts but even exhibit an unexpected upward trend, further confirming that global temporal metrics are ill-suited for evaluating localized restoration quality in object removal. Extended analyses can be found in
\ifarxiv
Sec.~\ref{sec:extended_fig5}
\else
Sec.~D.3
\fi
in the appendix.

\section{Proposed RC Metrics}

We introduce a unified local distribution matching framework in deep semantic feature space, named \textbf{Removal Coherence (RC)}, instantiated as two complementary metrics: \textbf{RC-S} for Spatial coherence within a frame, and \textbf{RC-T} for Temporal consistency across adjacent frames, as shown in Fig.~\ref{fig:method}.

\subsection{Unified Local Distribution Matching}

The key idea of RC is to evaluate object removal locally rather than relying on global feature aggregation. Given a restored image or video, we crop a local evaluation region around the target mask, extract semantic features, align the mask to the feature resolution, and compare local feature distributions using a sliding window.

\textbf{Local Cropping.}
Given an input mask $M$, we apply connected component analysis to obtain independent erased targets $\{m_k\}_{k=1}^{K}$. For each target $m_k$, we compute its bounding box $B_k$ and expand it outward by $1/3$ of its side length to include sufficient surrounding context. We then crop the restored image $I_{\text{out}}$ and the corresponding mask within the expanded box to obtain local pairs $(I_{\text{crop}}^{(k)}, M_{\text{crop}}^{(k)})$.

\textbf{Feature Extraction and Mask Alignment.}
For each cropped region, we use DINOv2~\cite{oquab2024dinov2} as the pretrained backbone $\Phi(\cdot)$ to extract a deep feature map
\begin{equation}
F^{(k)}=\Phi\!\left(I_{\text{crop}}^{(k)}\right)\in\mathbb{R}^{C\times H'\times W'},
\end{equation}
where $C$ is the channel dimension and $H' \times W'$ is the spatial resolution.
The cropped mask is downsampled to the same resolution to obtain the aligned binary mask $M'^{(k)}$.

\textbf{Window-wise Distribution Comparison.}
A $w \times w$ window is slid over the feature map. At each location $p$ with receptive field $\Omega_p$, we collect local feature sets and measure their discrepancy using the squared Maximum Mean Discrepancy (MMD):
\begin{equation}
\begin{split}
\mathrm{MMD}^2(X,Y)=
&\frac{1}{m^2}\sum_{i=1}^{m}\sum_{j=1}^{m}K(x_i,x_j)
+\frac{1}{n^2}\sum_{i=1}^{n}\sum_{j=1}^{n}K(y_i,y_j) \\
&-\frac{2}{mn}\sum_{i=1}^{m}\sum_{j=1}^{n}K(x_i,y_j),
\end{split}
\label{eq:mmd}
\end{equation}
where $X=\{x_i\}_{i=1}^{m}$ and $Y=\{y_j\}_{j=1}^{n}$ are two empirical feature sets, and $K(\cdot,\cdot)$ is a Gaussian RBF kernel.

\subsection{Spatial Metric: RC-S}

A high-quality removal result should exhibit local consistency between the restored content and its surrounding background. RC-S quantifies this spatial coherence by comparing the feature distributions of masked and unmasked regions within each local window.

For the $k$-th cropped target and window location $p$, we define the masked feature set:
\begin{equation}
\mathcal{X}_{\text{mask}}^{(k,p)}
=
\left\{
F^{(k)}(i)\mid i\in\Omega_p,\; M'^{(k)}(i)=1
\right\},
\end{equation}
and the local background feature set:
\begin{equation}
\mathcal{X}_{\text{bg}}^{(k,p)}
=
\left\{
F^{(k)}(i)\mid i\in\Omega_p,\; M'^{(k)}(i)=0
\right\},
\end{equation}
where $i$ indexes spatial locations on the feature map. The local spatial discrepancy is:
\begin{equation}
d_{\text{spatial}}^{(k,p)}
=
\mathrm{MMD}^2\!\left(
\mathcal{X}_{\text{mask}}^{(k,p)},
\mathcal{X}_{\text{bg}}^{(k,p)}
\right).
\end{equation}

When a window falls entirely within the mask such that $\mathcal{X}_{\text{bg}}^{(k,p)}=\emptyset$, all background features in the cropped view are used as the reference. Let $\mathcal{P}^{(k)}$ denote the set of valid windows intersecting the mask for the $k$-th target. RC-S is computed by averaging window-level discrepancies within each target and then across all targets:
\begin{equation}
\mathrm{RC\text{-}S}
=
\frac{1}{K}
\sum_{k=1}^{K}
\left(
\frac{1}{|\mathcal{P}^{(k)}|}
\sum_{p\in\mathcal{P}^{(k)}} d_{\text{spatial}}^{(k,p)}
\right).
\end{equation}
A lower RC-S value indicates better spatial coherence. For presentation, we apply an inverse normalization, defined as $\exp(-\text{RC-S}/ \tau)$ with $\tau=3$, so that higher values denote better quality.

\subsection{Temporal Metric: RC-T}

Beyond spatial coherence, video object removal additionally requires the restored regions to remain temporally stable across frames. RC-T follows the same local distribution matching pipeline as RC-S, but compares feature distributions of restored regions across adjacent frames rather than between masked and unmasked regions within a single frame.

For two adjacent frames $(I^{(t)}, M_t)$ and $(I^{(t+1)}, M_{t+1})$,  cropping each frame independently may introduce spatial misalignment. We therefore define a shared local evaluation region using the union of the two masks:
\begin{equation}
M_{\text{union}} = M_t \cup M_{t+1}.
\end{equation}
The bounding box of $M_{\text{union}}$ is expanded outward by one-third of its side length, and both frames and masks are synchronously cropped to obtain
$(I_{\text{crop}}^{(t)}, M_{\text{crop}}^{(t)})$ and
$(I_{\text{crop}}^{(t+1)}, M_{\text{crop}}^{(t+1)})$.

The corresponding feature maps are extracted as:
\begin{equation}
F^{(s)}=\Phi\!\left(I_{\text{crop}}^{(s)}\right)\in\mathbb{R}^{C\times H'\times W'},
\quad s\in\{t,t+1\},
\end{equation}
and the two cropped masks are downsampled to obtain aligned masks $M'^{(t)}$ and $M'^{(t+1)}$. To focus the evaluation on regions restored in both frames, we use their intersection:
\begin{equation}
M_{\cap}=M'^{(t)} \cap M'^{(t+1)}.
\end{equation}

For each window location $p$, we collect local features inside the shared restored region from the two frames:
\begin{equation}
\begin{aligned}
\mathcal{X}_{t}^{(p)}
&=
\left\{
F^{(t)}(i)\mid i\in\Omega_p,\; M_{\cap}(i)=1
\right\},\\
\mathcal{X}_{t+1}^{(p)}
&=
\left\{
F^{(t+1)}(i)\mid i\in\Omega_p,\; M_{\cap}(i)=1
\right\},
\end{aligned}
\end{equation}
and compute the local temporal discrepancy as:
\begin{equation}
d_{\text{temporal}}^{(t,p)}
=
\mathrm{MMD}^2\!\left(
\mathcal{X}_{t}^{(p)},
\mathcal{X}_{t+1}^{(p)}
\right).
\end{equation}

Let $\mathcal{P}^{(t)}$ denote the set of valid windows intersecting the shared restored region between frames $t$ and $t+1$. RC-T averages the local discrepancies over valid windows and then over all adjacent frame pairs:
\begin{equation}
\mathrm{RC\text{-}T}
=
\frac{1}{T-1}
\sum_{t=1}^{T-1}
\left(
\frac{1}{|\mathcal{P}^{(t)}|}
\sum_{p\in\mathcal{P}^{(t)}} d_{\text{temporal}}^{(t,p)}
\right).
\end{equation}
A lower RC-T value indicates higher temporal consistency of the restored regions across frames.

\section{PROVE-Bench}
\label{sec:bench}

We construct \textbf{PROVE-Bench}, comprising two complementary real-world subsets: \textbf{PROVE-M} and \textbf{PROVE-H}. PROVE-M is built from paired recordings augmented with simulated camera motion, providing aligned input video, mask, and target-free ground-truth video triplets for realistic yet controllable evaluation. In contrast, PROVE-H contains highly challenging videos without ground truth, designed to assess model robustness, generalization, and metric behavior under unconstrained conditions.

\subsection{PROVE-M: Motion-Augmented Real-World Paired Benchmark}
\label{PROVE-M}
\textbf{Real-World Paired Capture.}
For each scene, we record two consecutive videos using a tripod-mounted \textit{stationary camera} to form a paired sample. We first capture the input video $V_{in}$ containing the target object, then remove the object and immediately capture the corresponding target-free video $V_{gt}$. Both recordings are completed within two minutes to maintain consistent illumination, shadow direction, and scene layout. Dynamic background elements such as pedestrians or vehicles are carefully controlled to minimize unrelated temporal changes. Object masks are obtained using SAM3~\cite{carion2025sam3segmentconcepts} and manually refined frame by frame, yielding accurate per-frame annotations. Each PROVE-M sample thus consists of an input video, a mask video, and a target-free ground-truth video.

\textbf{Pairwise Quality Control.}
Despite controlled capture conditions, unavoidable acquisition factors may still
introduce imperfect video pairs. We apply a three-stage filtering pipeline to ensure data reliability. In the \textbf{first stage}, we assess pair quality based on mask consistency rather than raw RGB similarity: a consistency score is computed as the PSNR between a difference-based coarse mask $M_{diff}$ and the refined ground-truth mask $M_{gt}$, and only the top 40\% of candidates are retained. In the \textbf{second stage}, samples with severe background disturbances are removed by detecting large connected components outside the removal mask. \textbf{Finally}, the remaining videos are manually inspected, yielding 80 high-quality paired cases as the PROVE-M source set. More details can be found in
\ifarxiv
Sec.~\ref{sec:pairwise_control}
\else
Sec.~B.2
\fi
in the appendix.

\textbf{Motion Augmentation.}
To bridge the gap between stationary recordings and real user-captured videos, we augment the paired recordings with simulated camera motion via Ken Burns-style geometric transformations, including cropping, scaling, and translation, mimicking common camera behaviors such as handheld shake, push/pull zoom, and target-following motion. The same transformation is applied synchronously to the entire triplet $(V_{in}, V_{gt}, M_{gt})$, preserving strict frame-wise alignment after augmentation. Each video comprises 81 frames at 1080p resolution, and covers both landscape and portrait layouts. Since camera motion naturally amplifies temporal instability and boundary artifacts, PROVE-M poses substantially greater challenges for both removal models and evaluation metrics. As shown in
\ifarxiv
Sec.~\ref{sec:bench_result}
\else
Sec.~B.3
\fi
in the appendix, existing models consistently degrade under this dynamic setting, underscoring the necessity of motion-aware evaluation.

\subsection{PROVE-H: Harder Real-World Benchmark without Ground Truth}

To complement the paired setting of PROVE-M, we construct PROVE-H, a hard real-world benchmark \textit{without target-free ground truth}. PROVE-H targets challenging scenarios that are particularly relevant to object removal in unconstrained environments, including crowd scenes, dynamic backgrounds (e.g., flowing water, flames, rain, and snow), highly textured backgrounds (e.g., grasslands and deserts), complex reflections with intertwined side effects (e.g., multiple puddle reflections), and fast-motion scenes. These factors substantially increase removal difficulty and expose failure modes that are less visible in controlled paired settings. To faithfully reflect real-world deployment conditions, PROVE-H uses only SAM3-generated masks \textit{without manual refinement}, preserving the practical imperfections of automatic segmentation in unconstrained scenarios.

\begin{table}
    \centering
    \setlength{\tabcolsep}{2.5pt} 
    
    \caption{Comparison with existing open-source video object removal datasets. 
        \textbf{Real}: real-world source video. 
        \textbf{GT}: paired target-free ground truth. 
        \textbf{Sh.}: shadows. 
        \textbf{Ref.}: reflections. 
        \textbf{M.E.}: multiple simultaneous effects. 
        \textbf{D.A.}: disconnected associations. 
        \textbf{Crw.}: crowds and occlusions. 
        \textbf{Tex.}: textured backgrounds. 
        \textbf{Fst.}: fast motion.
    }
    \label{tab:benchmark_comparison}
    
    \resizebox{\linewidth}{!}{
    \begin{tabular}{l | c c | c c | c c c c c | c}
        \toprule
        \multirow{2}{*}{\textbf{Dataset}} & \multicolumn{2}{c|}{\textbf{Base}} & \multicolumn{2}{c|}{\textbf{Basic}} & \multicolumn{5}{c|}{\textbf{Advanced Challenges}} & \multirow{2}{*}{\textbf{\#}} \\
        \cmidrule(lr){2-3} \cmidrule(lr){4-5} \cmidrule(lr){6-10}
        & \textbf{Real} & \textbf{GT} & \textbf{Sh.} & \textbf{Ref.} & \textbf{M.E.} & \textbf{D.A.} & \textbf{Crw.} & \textbf{Tex.} & \textbf{Fst.} & \\
        \midrule
        DAVIS~\cite{pont20172017}           & \cmark & \xmark & \cmark & \cmark & \xmark & \xmark & \xmark & \cmark & \cmark & 90 \\
        Movies~\cite{lin2023omnimatterf}    & \xmark & \cmark & \cmark & \cmark & \xmark & \xmark & \xmark & \xmark & \cmark & 5 \\
        Kubric~\cite{wu2022d} & \xmark & \cmark & \cmark & \xmark & \xmark & \cmark & \xmark & \xmark & \xmark & 5 \\
        GenProp~\cite{liu2025generative} & \cmark & \xmark & \cmark & \cmark & \xmark & \xmark & \xmark & \xmark & \xmark & 15 \\
        ROSE-Bench~\cite{miao2025rose}  & \xmark & \cmark & \cmark & \cmark & \xmark & \cmark & \xmark & \xmark & \xmark & 60 \\
        \midrule
        \rowcolor{green!10} 
        \textbf{PROVE-M (Ours)} & \cmark & \cmark & \cmark & \cmark & \cmark & \cmark & \xmark & \xmark & \cmark & 80 \\
        \rowcolor{green!10} 
        \textbf{PROVE-H (Ours)} & \cmark & \xmark & \cmark & \cmark & \cmark & \cmark & \cmark & \cmark & \cmark & 100 \\
        \bottomrule
    \end{tabular}
    }
\end{table}

\begin{figure}
    \centering
    \begin{subfigure}{\linewidth}
        \centering
        \includegraphics[width=\linewidth]{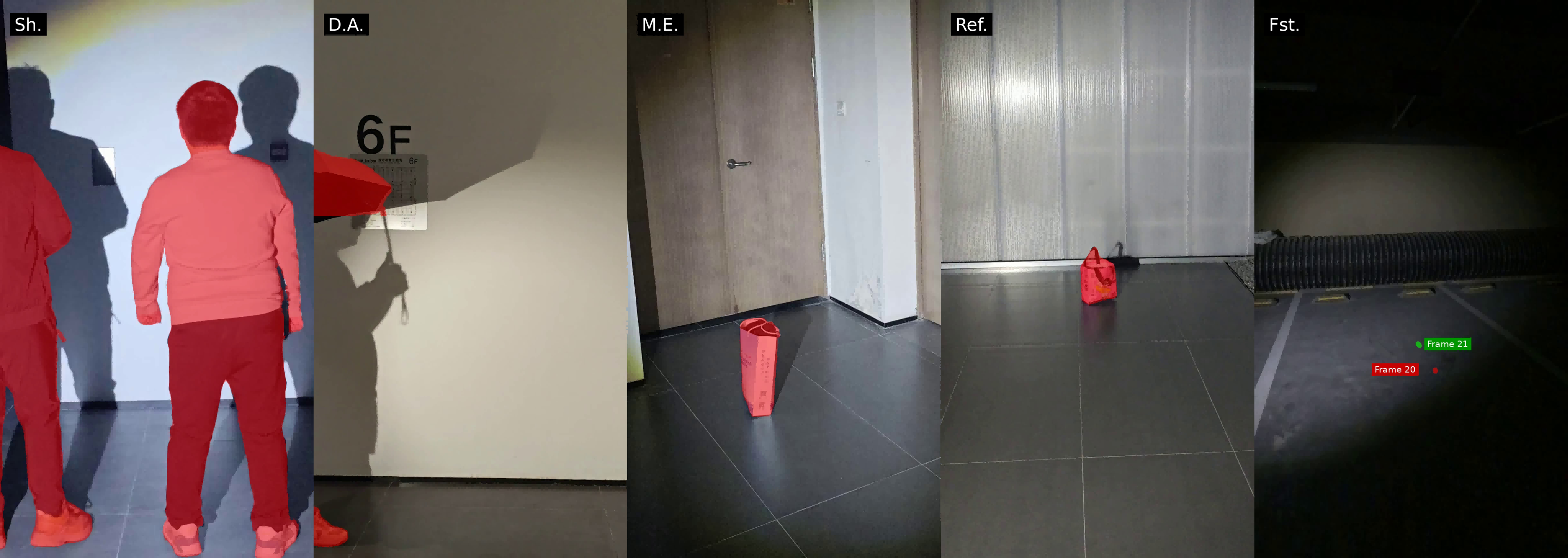}
        \caption{PROVE-M}
        \label{fig:m_concat_vis}
    \end{subfigure}

    \begin{subfigure}{\linewidth}
        \centering
        \includegraphics[width=\linewidth]{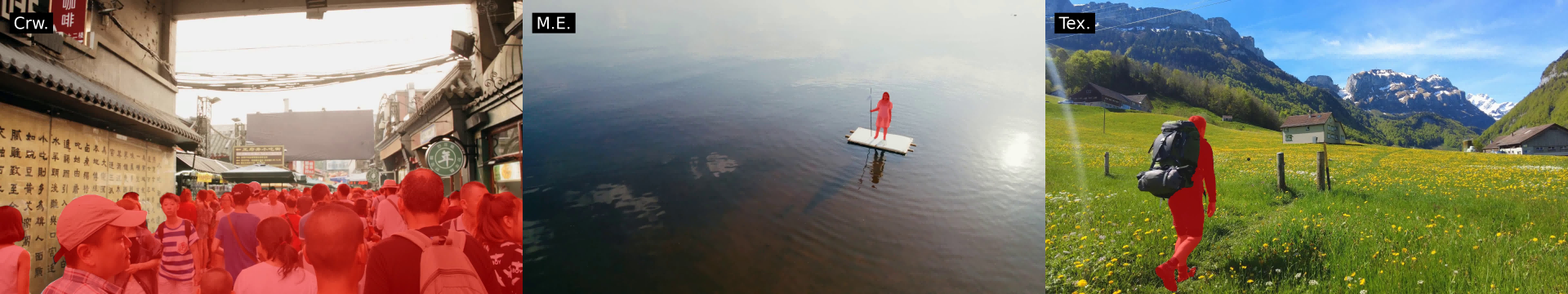}
        \caption{PROVE-H}
        \label{fig:h_concat_vis}
    \end{subfigure}

    \caption{Sample frames from the proposed PROVE-Bench.}
    \label{fig:prove_sample}
\end{figure}

\subsection{Benchmark Statistics and Comparison}

As shown in Table~\ref{tab:benchmark_comparison}, existing benchmarks face a fundamental realism--evaluability dilemma: real-world datasets such as DAVIS lack paired reference videos, while datasets with paired GT such as ROSE-Bench are synthetic. PROVE-M mitigates this tension by combining aligned real-world references with synchronized motion augmentation. Beyond this, while prior benchmarks adequately cover basic physical effects, they fall short on advanced challenges. PROVE-M captures intricate physical entanglements — including multiple simultaneous effects (M.E.) and disconnected associations (D.A.) — with full paired evaluability, while PROVE-H serves as an unconstrained stress test targeting extreme spatiotemporal dynamics such as dense crowds (Crw.) and highly textured backgrounds (Tex.).
Figure~\ref{fig:prove_sample} shows some examples from the PROVE-M and PROVE-H datasets. The full PROVE-Bench dataset and detailed dataset statistics are provided in the appendix (
\ifarxiv
Sec.~\ref{sec:dataset_statis}
\else
Sec.~B.1
\fi
).

\section{Experiments}
\label{sec:experiments}

\subsection{Experimental Setup}
\label{sec:setup}

\paragraph{\textbf{Datasets}.}
We conduct experiments across diverse scenarios encompassing both synthetic and real-world data, with and without paired target-free ground truth.

For \textbf{image datasets}, we use RORD-Val~\cite{sagong2022rord} and OBER-Wild~\cite{zhao2026objectclear}. RORD-Val is a widely used object removal benchmark with paired target-free GT; we adopt the 343 samples with manually refined masks~\cite{zhao2026objectclear}. OBER-Wild is a no-reference dataset of 300 real-world images featuring complex physical effects such as shadows and reflections.

For \textbf{video datasets}, we use DAVIS~\cite{pont20172017}, ROSE-Bench~\cite{miao2025rose}, and the proposed PROVE-Bench. DAVIS contains 90 real-world video sequences without paired GT. ROSE-Bench is a synthetic dataset rendered with Unreal Engine, comprising 60 videos across six categories of physical side effects. PROVE-Bench consists of PROVE-M (80 motion-augmented real-world paired videos) and PROVE-H (100 challenging real-world videos without ground truth).

\paragraph{\textbf{Evaluated Methods}.}
For \textbf{image object removal}, we evaluate {LaMa}~\cite{suvorov2022resolution} (non-diffusion), {SmartEraser}~\cite{jiang2025smarteraser}, {OmniEraser}~\cite{wei2025omnieraser}, and {ObjectClear}~\cite{zhao2026objectclear}. OmniEraser and ObjectClear are specifically optimized for handling shadows and reflections.
For \textbf{video object removal}, we evaluate {FGT}~\cite{zhang2022flow} (non-diffusion), {Minimax}~\cite{zi2025minimax}, {ROSE}~\cite{miao2025rose}, and {GenOmni}~\cite{lee2025generative}; the latter two are capable of restoring complex spatiotemporal side effects.

\paragraph{\textbf{Evaluation Metrics}.}
For full-reference evaluation, beyond standard global metrics (PSNR, SSIM, LPIPS), we
introduce two sets of localized derivatives — mask-only ($m$-PSNR/SSIM/LPIPS) and
background-only ($bg$-PSNR/SSIM/LPIPS) — to separately assess regional performance. For no-reference evaluation, we compare proposed {RC-S} against {ReMOVE}~\cite{chandrasekar2024remove} and
{CFD}~\cite{yu2025omnipaint}.

\paragraph{\textbf{Human Study Protocol}.}
We conduct a subjective preference study with 20 participants. For each image or video, participants are shown the masked input, the GT (when available), and the outputs of four models in randomized order. They are then asked to rank the results by overall removal quality and contextual coherence. Rankings are aggregated using Borda Count~\cite{emerson2013original}, assigning scores of 3, 2, 1, and 0 for 1st through 4th place.

\paragraph{\textbf{Correlation Computation}.}
To quantify alignment with human perception, we compute Kendall's Tau ($\tau$) and
Spearman correlation ($\rho$) between metric-induced rankings and aggregated human
scores. Furthermore, supplementary experiments using GPT-4o as an auxiliary evaluator are detailed in the appendix (
\ifarxiv
Sec.~\ref{sec:eval_gpt}
\else
Sec.~C.5
\fi
).

\subsection{Human Correlation Analysis of RC-S}
\label{sec:results}

\begin{table*}[t]
\centering
\caption{\textbf{Correlation with human rankings.}
We report Kendall's $\tau$ and Average Spearman correlation $\rho$ between metric-induced rankings and aggregated human rankings.
Bold and underline indicate the best and second-best values in each column, respectively.
``--'' indicates the metric is not applicable or unavailable for that benchmark.
AVG is computed over available benchmarks only.}
\label{tab:human_corr}
\footnotesize 
\setlength{\tabcolsep}{4.5pt} 
\renewcommand{\arraystretch}{1.1}
\begin{tabular}{l ccccccc c ccccccc}
\toprule[1.2pt]
& \multicolumn{7}{c}{\textbf{Kendall's $\tau$}} & & \multicolumn{7}{c}{\textbf{Avg. Spearman $\rho$}} \\
\cmidrule{2-8} \cmidrule{10-16}
\textbf{Metric}
& RORD & OBER-Wild & DAVIS & ROSE & PROVE-M & PROVE-H & AVG
&
& RORD & OBER-Wild & DAVIS & ROSE & PROVE-M & PROVE-H & AVG \\
\midrule
PSNR         &  0.01 & --    & --    &  0.36 & 0.38 & -- &  0.25 &
             &  0.02 & --    & --    &  0.44 & 0.45 & -- &  0.30 \\
SSIM         & -0.22 & --    & --    &  0.11 & 0.43 & -- &  0.11 &
             & -0.31 & --    & --    &  0.11 & 0.46 & -- &  0.09 \\
LPIPS        & -0.23 & --    & --    &  0.24 & 0.33 & -- &  0.12 &
             & -0.28 & --    & --    &  0.28 & 0.37 & -- &  0.13 \\
\midrule

\rowcolor{gray!8} m-PSNR       & -0.16 & --    & --    &  0.48 & 0.53 & -- &  0.29 &
\cellcolor{white} & -0.17 & --    & --    &  0.57 & 0.62 & -- &  0.34 \\
\rowcolor{gray!8} m-SSIM       &  0.08 & --    & --    &  0.53 & \underline{0.68} & -- &  0.43 &
\cellcolor{white} &  0.10 & --    & --    &  0.62 & \underline{0.73} & -- &  0.48 \\
\rowcolor{gray!8} m-LPIPS      &  \underline {0.19} & --    & --    & \textbf{0.68} & \underline{0.68} & -- & \underline{0.52} &
\cellcolor{white} &  \underline{0.24} & --    & --    & \textbf{0.75} & \textbf{0.75} & -- & \underline{0.58} \\
\midrule
bg-PSNR      & -0.13 & -0.42 & -0.52 &  0.23 & 0.26 & -0.66 & -0.21 &
             & -0.16 & -0.53 & -0.59 &  0.29 & 0.32 & -0.73 & -0.23 \\
bg-SSIM      & -0.26 & -0.15 & -0.57 &  0.01 & 0.27 & -0.74 & -0.24 &
             & -0.34 & -0.23 & -0.65 &  0.00 & 0.31 & -0.80 & -0.29 \\
bg-LPIPS     & -0.33 & -0.33 & -0.63 & -0.03 & 0.06 & -0.73 & -0.33 &
             & -0.40 & -0.43 & -0.69 & -0.01 & 0.11 & -0.80 & -0.37 \\
\midrule
\rowcolor{gray!8} ReMOVE  &  0.06 &  \underline{0.54} &  0.15 &  0.21 & 0.33 &  \underline{0.23} &  0.26 &
\cellcolor{white} &  {0.08} &  \underline{0.61} &  0.16 &  0.24 & 0.36 &  \underline{0.27} &  0.29 \\
\rowcolor{gray!8} CFD  & -0.04 &  0.40 &  \underline{0.21} &  0.03 & 0.24 &  0.12 &  0.16 &
\cellcolor{white} & -0.05 &  0.47 &  \underline{0.25} &  0.04 & 0.26 &  0.14 &  0.18 \\
\midrule

\rowcolor{cyan!6} \textbf{RC-S} & \textbf{0.31} & \textbf{0.57} & \textbf{0.60} & \underline{0.61} & \textbf{0.70} & \textbf{0.76} & \textbf{0.59} &
\cellcolor{white} & \textbf{0.39} & \textbf{0.66} & \textbf{0.68} & \underline{0.69} & \textbf{0.75} & \textbf{0.82} & \textbf{0.66} \\
\bottomrule[1.2pt]
\end{tabular}
\end{table*}

Table~\ref{tab:human_corr} presents the full correlation analysis of RC-S, from which we draw the following key observations.

{\textbf{The Illusion of Full-Reference Metrics}.}
Global FR metrics conflate target erasure quality with background fidelity, rewarding methods that preserve non-masked regions even when the erased area is visually poor — which explains why background-only variants exhibit consistently negative correlations across most benchmarks. Mask-only FR metrics are more task-relevant but remain heavily dependent on GT quality: they perform reasonably on ROSE-Bench and PROVE-M but generalize poorly to RORD. More critically, FR metrics cannot be applied in no-GT settings such as OBER-Wild, DAVIS, and PROVE-H, fundamentally limiting their utility in realistic evaluation scenarios.

{\textbf{Vulnerabilities of Existing No-Reference Metrics}.}
Regarding No-Reference (NR) evaluation, existing metrics such as CFD and ReMOVE demonstrate suboptimal performance, with CFD being particularly ineffective. As analyzed in Sec. \ref{sec:nr_limitations}, both metrics suffer from the issue of ``averaging differences,'' which dilutes localized artifacts. Furthermore, CFD's reliance on SAM makes it highly susceptible to over-segmentation errors, while ReMOVE lacks the necessary sensitivity to accurately evaluate multi-object erasure scenarios.

Notably, RC-S achieves the best average correlation and ranks first on five of the six benchmarks under both Kendall's $\tau$ and Spearman's $\rho$.

\subsection{Validation of RC-T}
Validating temporal metrics via direct human ranking is inherently challenging, because subtle localized temporal artifacts in natural videos are difficult to rank reliably by pairwise human preference. Therefore, as stated in Sec.~\ref{sec:lim_tm}, we adopt a sensitivity-based validation protocol. Using high-quality removal results on DAVIS as clean baselines, we introduce two types of controlled temporal corruption of increasing severity — Random Drop and Random Replace — and measure each metric's response. A well-calibrated temporal metric should exhibit monotonically degrading scores as corruption intensity increases.

As shown in Fig.~\ref{fig:both_datasets} (also
\ifarxiv
Fig.~\ref{fig:rct_suppl}
\else
Fig.~20
\fi
in the appendix), RC-T responds sensitively and monotonically to both corruption types, whereas TC and TF remain largely insensitive and even improve under certain conditions. This confirms that RC-T's localized distribution matching captures temporal artifacts that global frame-level metrics systematically miss.

\subsection{Ablation Study}
\label{sec:ablation}

\begin{table}[t]
\centering
\caption{\textbf{Ablation study of RC-S.} Correlation with human rankings measured by Kendall's $\tau$.}
\label{tab:ablation_rcs}
\resizebox{\linewidth}{!}{
\begin{tabular}{lccccccc}
\toprule
& \multicolumn{7}{c}{\textbf{Kendall's $\tau$}} \\
\cmidrule(lr){2-8}
\textbf{Variant} & RORD & OBER-Wild & DAVIS & ROSE & PROVE-M & PROVE-H & AVG \\
\midrule
RC-S (w/o window) & 0.18 & 0.44 & 0.53 & 0.51 & 0.55 & 0.64 & 0.48 \\
RC-S (w/ cosine)  & 0.26 & 0.52 & 0.58 & 0.50 & 0.63 & 0.65 & 0.52 \\
RC-S (SAM)        & 0.27 & \textbf{0.63} & 0.41 & 0.43 & 0.43 & 0.48 & 0.44 \\
RC-S (DINOv3) & 0.30 & 0.53 & 0.54 & 0.49 & 0.62 & 0.59 & 0.51\\
RC-S (DINOv2)     & \textbf{0.31} & 0.57 & \textbf{0.60} & \textbf{0.61} & \textbf{0.70} & \textbf{0.76} & \textbf{0.59} \\
\bottomrule
\end{tabular}}
\end{table}

We ablate RC-S along three design dimensions: feature representation, locality of comparison, and discrepancy metric. Table \ref{tab:ablation_rcs} compares the default RC-S against three representative variants.

{\textbf{Feature extractor (RC-S w/ SAM and DINOv3~\cite{simeoni2025dinov3})}:} We compare DINOv2 with two alternative feature extractors, SAM and DINOv3. While SAM is effective for boundary delineation, DINOv2 provides a more perceptually sensitive feature space for assessing fine-grained local coherence. This is consistent with prior findings on DINOv2's stronger alignment with low-level human visual characteristics~\cite{cai2025computer}, and with our Fourier-domain sensitivity analysis (see
\ifarxiv
Sec.~\ref{sec:supp_fourier_analysis}
\else
Sec.~A.1
\fi
). We further evaluate DINOv3 as an additional backbone. Although it remains competitive, its overall performance is still lower than DINOv2. Notably, both the SAM and DINOv3 variants still outperform several existing metrics in Table~\ref{tab:human_corr}, demonstrating the robustness of our local distribution matching strategy across different feature backbones.

{\textbf{Local Assessment (RC-S w/o window)}:} Removing the sliding window consistently degrades performance. Global aggregation tends to dilute localized artifacts, whereas window-based comparison exposes regional inconsistency more explicitly, better matching human visual inspection.

{\textbf{Distance metric (RC-S w/ cosine)}:} Replacing MMD with cosine similarity also reduces correlation. Cosine similarity captures only first-order directional similarity and is less sensitive to distributional degradation, whereas MMD more accurately measures the local distribution shift between the restored region and its surrounding context.

\begin{figure}[t]
  \centering
  \includegraphics[width=\linewidth]{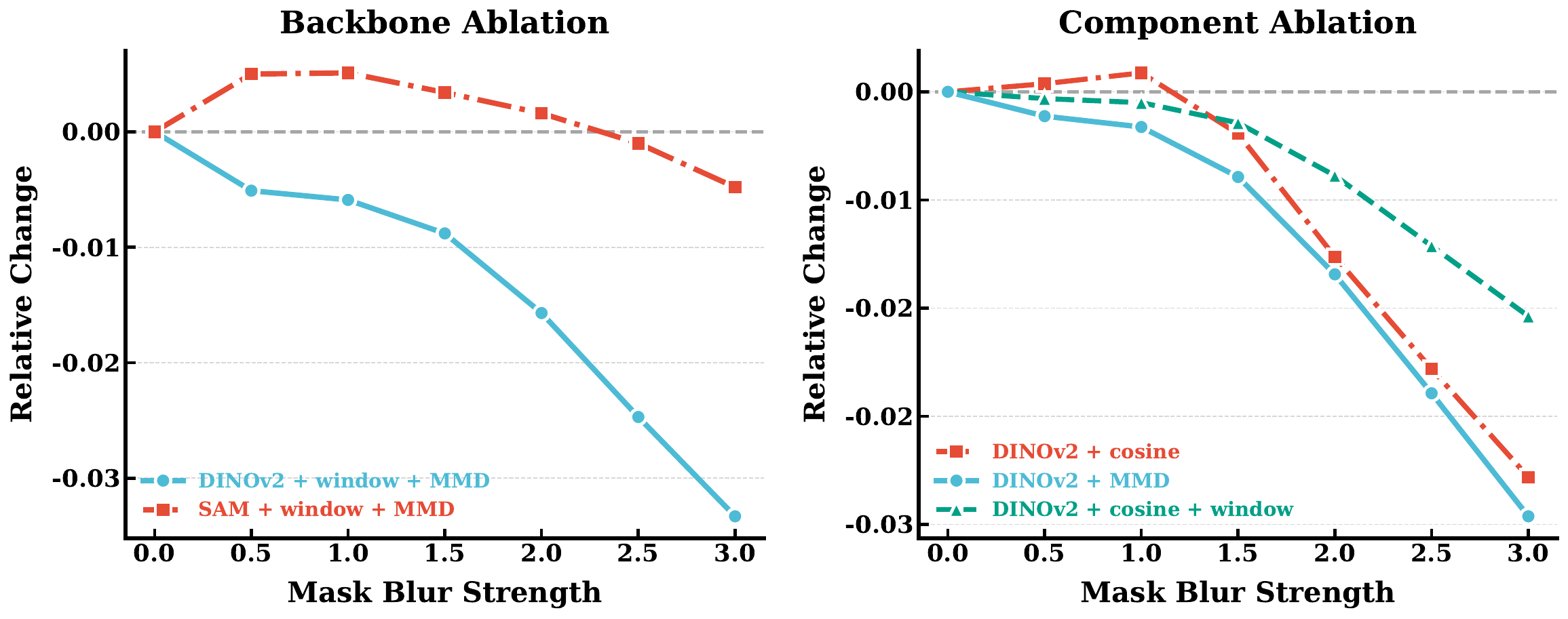}
  \caption{Ablation of RC-S under increasing mask blur. Left: backbone ablation. Right: component ablation. Relative changes are computed with respect to the zero-blur baseline.}
  \label{fig:blur_ablation}
\end{figure}

Beyond overall ranking correlation, we further examine which design choices confer robustness against the ``Blur is Clean'' bias. A human-aligned metric should produce lower scores as the blur intensity inside the masked region increases. As shown in Fig.~\ref{fig:blur_ablation}, we observe that \textbf{DINOv2 + window + MMD} follows this desired monotonic trend, whereas \textbf{SAM + window + MMD} does not, indicating that resistance to blur first depends on a perceptually meaningful feature space. Under the same DINOv2 features, \textbf{DINOv2 + cosine} fails to preserve the correct trend, while \textbf{DINOv2 + cosine + window} restores it, showing that localized comparison is crucial for exposing blur artifacts that would otherwise be diluted by global aggregation. Moreover, \textbf{DINOv2 + MMD} also exhibits the desired monotonic behavior, suggesting that MMD is inherently more sensitive to blur-induced distribution shifts than cosine similarity.

In summary, DINOv2, localized windowing, and MMD play complementary roles in representation, locality, and discrepancy measurement, respectively. Their synergy not only improves alignment with human rankings, but also makes RC-S more robust. Additional ablation studies on kernel parameters and window sizes are provided in the appendix (
\ifarxiv
Sec.~\ref{Additional Ablation}
\else
Sec.~D.5
\fi
). We also include an RC-T ablation in
\ifarxiv
Sec.~\ref{RC-T Ablation}
\else
Sec.~D.4
\fi
, showing that cropping is necessary to preserve sensitivity to localized temporal corruption.

\section{Discussion}
\label{sec:conclusion}

Object removal is inherently a one-to-many problem, yet existing metrics---whether full-reference, no-reference, or global temporal---fail to capture the perceptual quality of localized edits. Our results show that task-aligned evaluation should assess spatial coherence with surrounding context and temporal stability across frames, rather than fidelity to a single reference. To this end, we propose \textbf{RC} and \textbf{PROVE-Bench}, forming the \textbf{PROVE} framework: RC-S and RC-T assess region-aware spatial and temporal coherence, while PROVE-Bench combines paired real-world videos with challenging ground-truth-free scenarios for rigorous evaluation. Extensive experiments confirm that RC aligns better with human judgments than existing protocols. We hope PROVE serves as a practical foundation for standardized benchmarking of object removal methods.

\section*{Acknowledgements}
This work uses the OBER-Wild dataset (\url{https://github.com/zjx0101/ObjectClear}), licensed under NTU S-Lab License 1.0. The ROSE dataset (\url{https://huggingface.co/datasets/Kunbyte/ROSE-Dataset}), licensed under CC BY-NC 4.0. The authors confirm that all uses of the above resources are strictly for academic research purposes and not for any commercial application.

\bibliographystyle{ACM-Reference-Format}
\balance
\bibliography{main}

\ifarxiv
  \appendix
  \clearpage
\onecolumn
\section*{Appendix}
\label{sec:supple}

This appendix provides additional analyses and implementation details that complement the main paper. Specifically, Sec.~\ref{sec:Deep Dive into Evaluation Metrics} presents a deeper analysis of existing evaluation metrics and the design choices of our metric, including frequency-domain sensitivity, the ``Blur is Clean'' bias, and further discussions on CFD, ReMOVE, and TokSim. Sec.~\ref{sec:supple_bench_details} provides comprehensive details of PROVE-Bench, including dataset statistics, construction pipeline, and full benchmark results. Sec.~\ref{sec:evaluation details} further examines the robustness of both human and automated evaluation. Sec.~\ref{sec:Extended Experimental Results} reports extended experimental results, including runtime, additional visualizations, and ablation studies. Finally, Sec.~\ref{sec:Extended Discussion on Limitations} discusses the limitations of our method and benchmark.

\suppressfloats[t]
\section{A Deep Dive into Evaluation Metrics}
\label{sec:Deep Dive into Evaluation Metrics}
\subsection{Choice of DINOv2: Frequency-Domain Analysis and Feature Sensitivity}
\label{sec:supp_fourier_analysis}

To better understand the spectral characteristics of degraded removal results and the suitability of different feature backbones, we conduct two complementary frequency-domain analyses. We first analyze the spectral signatures of reduced-step inference and Gaussian blur, and then evaluate how different pretrained visual encoders respond to frequency perturbations.

\paragraph{Spectral signatures of reduced-step inference and blur.}
For each paired frame, we first convert the image to grayscale and compute its 2D Fast Fourier Transform (FFT). To reduce the dynamic range of spectral magnitudes, we use the centered log-magnitude spectrum for analysis. Given two paired images \(I_a\) and \(I_b\), we compute their log-magnitude spectra \(S_a\) and \(S_b\), and obtain a spectral difference map by pixel-wise subtraction:
\begin{equation}
D = S_a - S_b.
\end{equation}
We then average the difference maps over all paired frames to obtain a global frequency-difference map.

The results reveal clear spectral shifts under degradation. In both Fig.~\ref{fig:fourier_diff}(a) and Fig.~\ref{fig:fourier_diff}(b), low-step outputs exhibit stronger high-frequency responses, and the four broad red clusters suggest artifact-like residuals rather than faithful fine-detail reconstruction. This observation is also consistent with visual inspection, where erased regions under low-step inference often appear blurrier and less stable. By contrast, higher-step results retain more coherent fine details. Figure~\ref{fig:fourier_diff}(c) shows a different pattern: Gaussian blur produces a more regular low-pass spectral shift, corresponding to the systematic suppression of high-frequency texture details. Together, these results indicate that degraded removal outputs are closely associated with frequency-domain distortions.

\begin{figure}[!htbp]
  \centering
  \includegraphics[width=\linewidth]{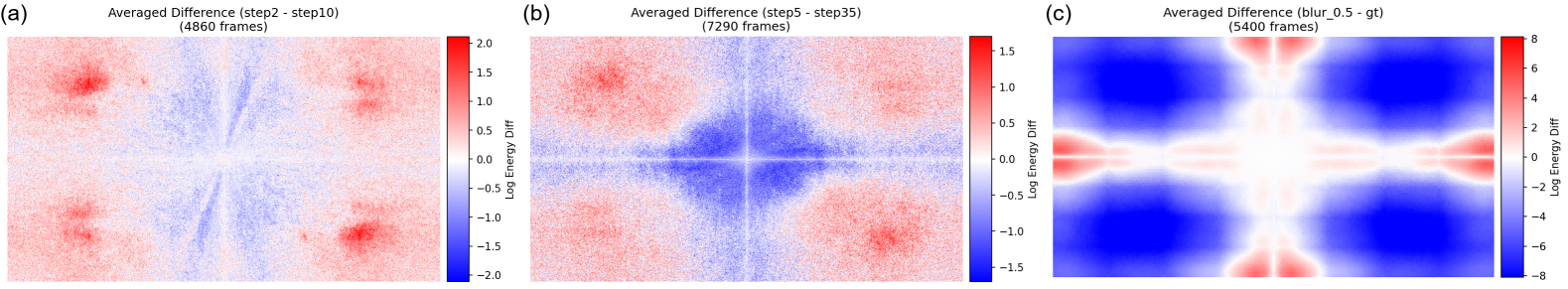}
  \caption{Frequency-domain comparison of reduced-step inference and blur.
(a) Minimax, step2 vs.\ step10 on ROSE-Bench.
(b) ROSE, step5 vs.\ step35 on DAVIS.
(c) Ground truth vs.\ Gaussian blur (\(\sigma=0.5\)) inside the mask on ROSE-Bench.}
  \label{fig:fourier_diff}
\end{figure}

\paragraph{Fourier error sensitivity of different backbones.}
We next evaluate how different pretrained visual encoders respond to such spectral perturbations. On DAVIS, for each clean frame \(I\), we inject a single 2D Fourier basis perturbation at spatial frequency \((u,v)\). The frequency space is uniformly sampled on a \(31 \times 31\) grid centered at the zero-frequency component. To ensure fair comparison across frequencies, each perturbation is first \(\ell_2\)-normalized and then scaled to a fixed magnitude \(\epsilon = 4.0\).

Let \(I_{u,v} = I + \delta_{u,v}\) denote the perturbed frame at frequency \((u,v)\). For each backbone \(f(\cdot)\), we extract the global image features of the clean and perturbed frames, and define the sensitivity at frequency \((u,v)\) as
\begin{equation}
s(u,v; I) = \left\| f(I + \delta_{u,v}) - f(I) \right\|_2 .
\end{equation}
We average the sensitivity values over all frames in DAVIS to obtain a frequency-response heatmap.

As shown in Fig.~\ref{fig:fourier_heatmap}, DINOv2~\cite{oquab2024dinov2} demonstrates a significantly broader and more intense sensitivity across the frequency spectrum compared to SAM~\cite{kirillov2023segment} and DINOv3~\cite{simeoni2025dinov3}. Quantitatively, the error magnitude of DINOv2 (max = $5.569$, mean = $3.824$) is drastically larger than that of SAM (max = $1.148$, mean = $0.935$) and DINOv3 (max = $0.841$, mean = $0.593$). While DINOv3 maintains an extremely robust feature space against high-frequency perturbations, DINOv2's feature representations are acutely vulnerable to them.

Together, these analyses fundamentally justify our design choices for RC-S and RC-T. We \textbf{adopt DINOv2 as the feature backbone because it exhibits substantially greater sensitivity to perturbations than SAM and DINOv3}. This makes DINOv2 better suited for detecting subtle local degradations, such as blur, texture loss, and artifact-like residuals, in object removal evaluation.

\begin{figure}[!htbp]
  \centering
  \includegraphics[width=\linewidth]{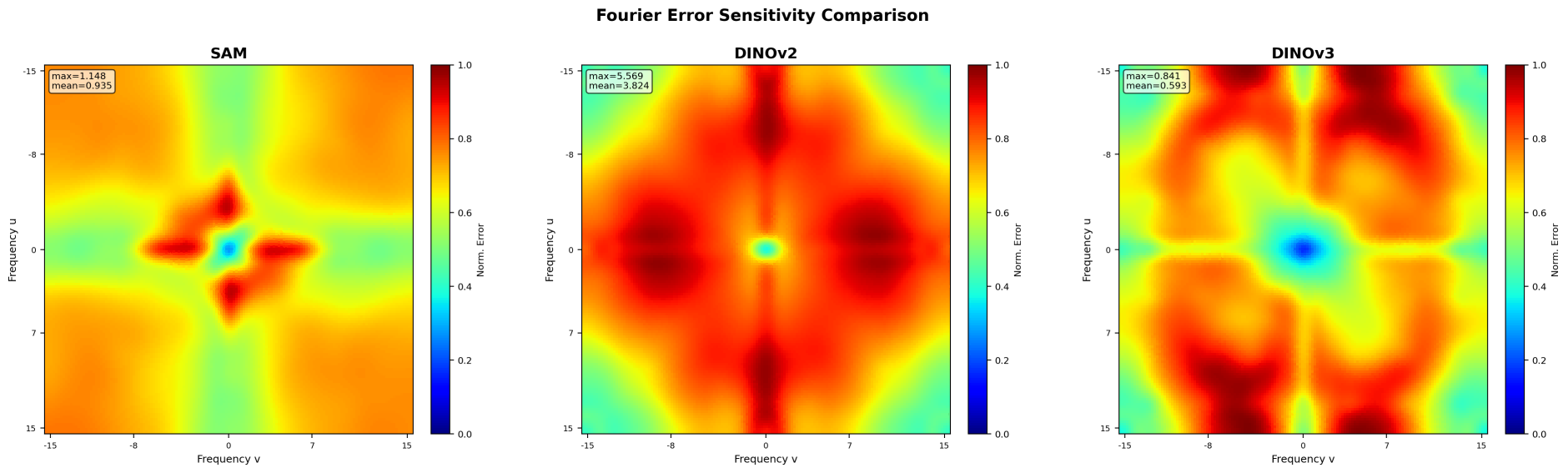}
  \caption{Fourier sensitivity of different backbones. DINOv2 shows broader and stronger responses to frequency perturbations than SAM and DINOv3, indicating higher sensitivity to spectral distortions.}
  \label{fig:fourier_heatmap}
\end{figure}

\subsection{Visualizing the ``Blur is Clean'' Bias}
As a visual supplement to the controlled blur experiment in Sec.~\ref{sec:nr_limitations}, we provide an example in Fig.~\ref{fig:blur_clean_vis}. Starting from the ground-truth frame, we progressively apply Gaussian blur with increasing strength only inside the mask region, while keeping the unmasked area unchanged. From left to right, the figure shows the mask, the ground truth, and two blurred variants (\textit{blur\_1} and \textit{blur\_3}). As the blur strength increases, the masked region becomes increasingly smoother, and fine textures and structural details are gradually suppressed. Although these blurred results are clearly worse perceptually, existing no-reference metrics may still assign them higher scores, because the blurred region becomes more homogeneous and thus appears more similar to the surrounding background in feature space. This visualization provides intuitive evidence for the ``Blur is Clean'' bias discussed in Sec.~\ref{sec:nr_limitations}.

\begin{figure}[t]
  \centering
  \includegraphics[width=\linewidth]{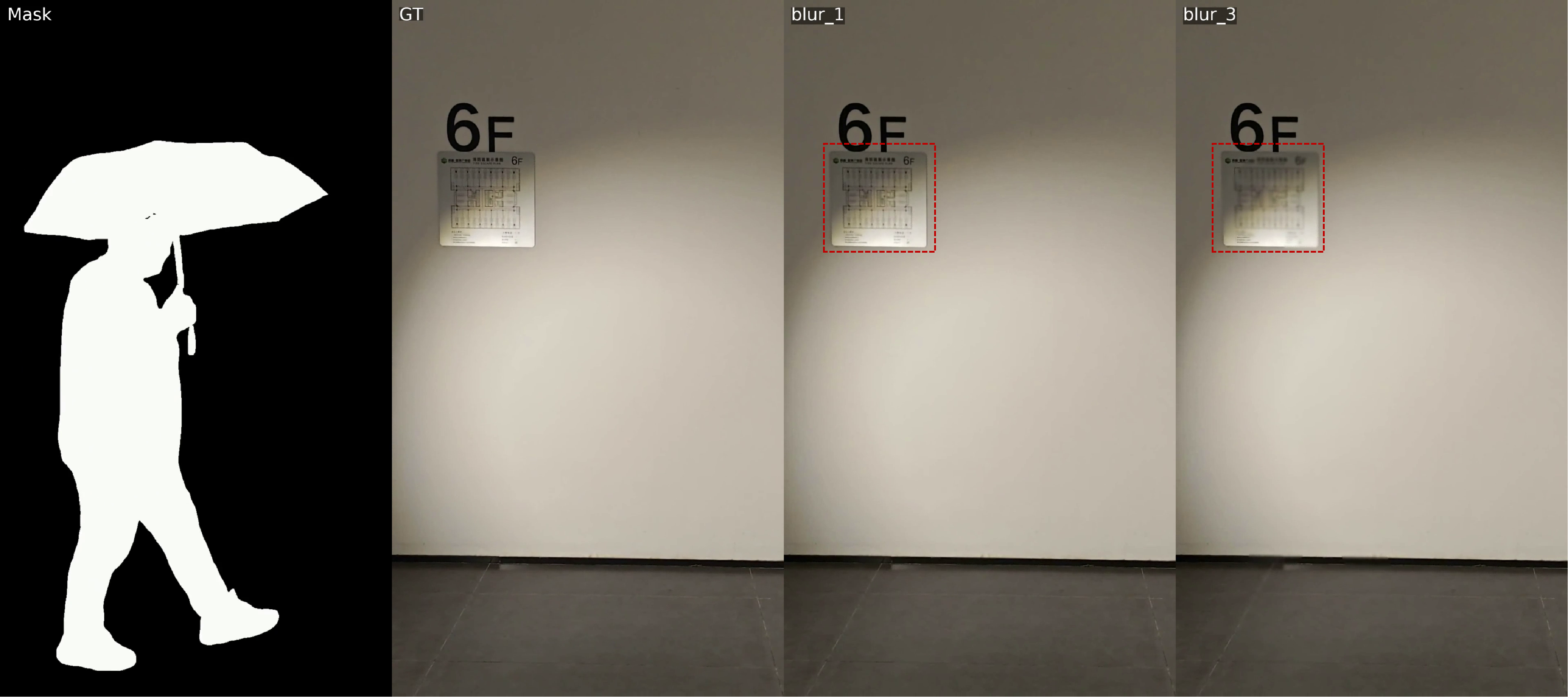}
  \caption{Visual supplement to the ``Blur is Clean'' bias in Sec.~\ref{sec:nr_limitations}. From left to right: mask, ground truth, and two blurred variants (\textit{blur\_1}, \textit{blur\_3}), where Gaussian blur with progressively increasing strength is applied only inside the mask region.}
  \label{fig:blur_clean_vis}
\end{figure}

\subsection{Further Analysis of Existing NR Metrics}
\label{sec:CFD}
\paragraph{\textbf{CFD}}
CFD~\cite{yu2025omnipaint} exhibits two additional issues. First, its \textit{context coherence} term may assign a better score to the original image than to the target-free ground truth, even when the hallucination penalty is zero. We attribute this mainly to the \texttt{[CLS]} token in DINOv2~\cite{oquab2024dinov2}, whose global semantic aggregation suppresses fine-grained local differences. As shown in Fig.~\ref{fig:cfd_cls}, removing \texttt{[CLS]} alleviates this abnormal behavior.

\begin{figure}[!htbp]
  \centering
  \includegraphics[width=0.8\linewidth]{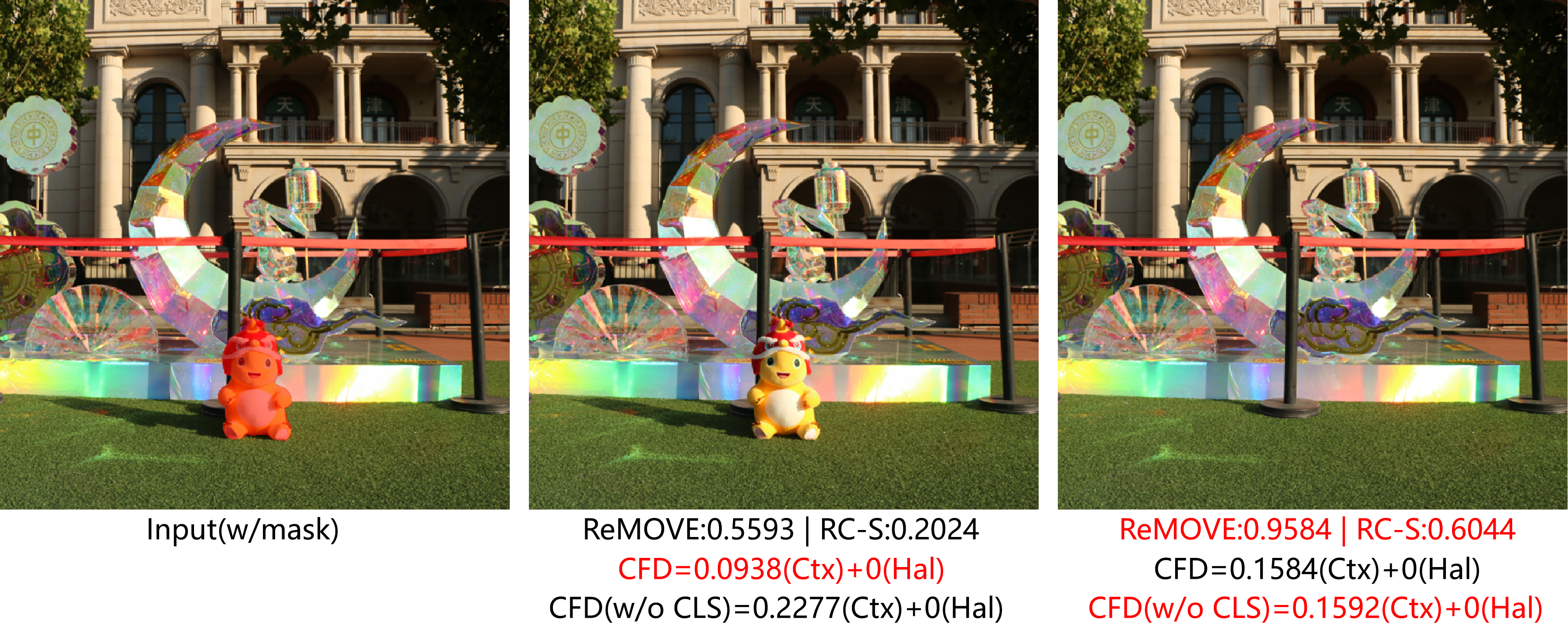}
  \caption{Failure case of the CFD context term caused by the \texttt{[CLS]} token. \textcolor{red}{Red} indicates the result preferred by the metric. Removing \texttt{[CLS]} alleviates this issue.}
  \label{fig:cfd_cls}
\end{figure}

\begin{figure}[!htbp]
  \centering
  \includegraphics[width=0.8\linewidth]{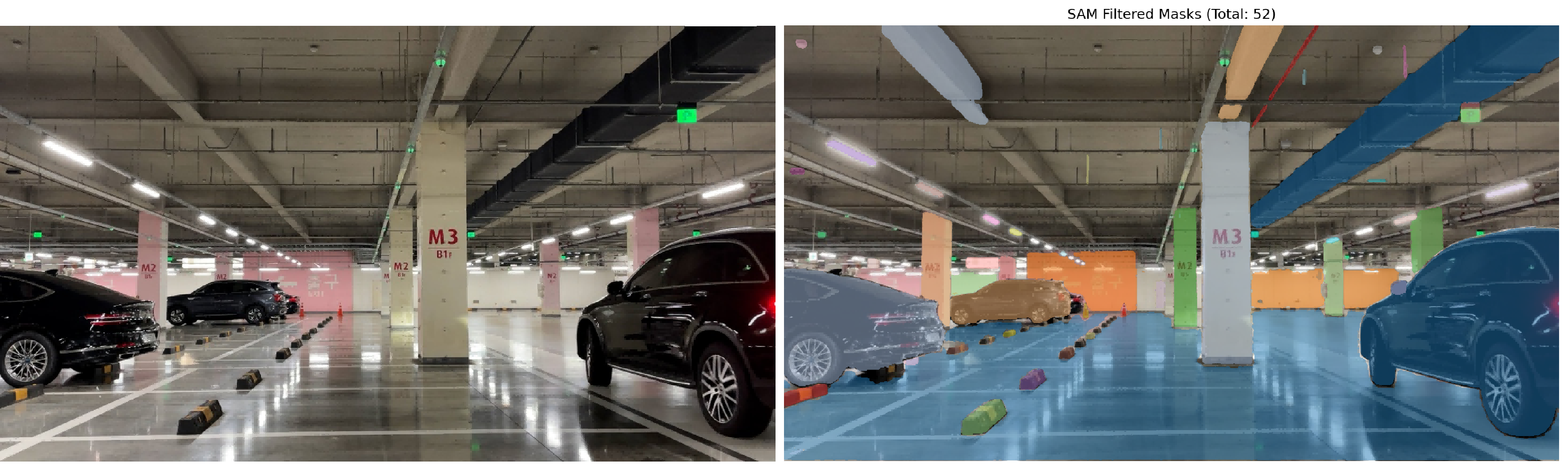}
  \caption{Over-segmentation of SAM in a complex scene. }
  \label{fig:cfd_sam}
\end{figure}

Second, CFD relies on SAM-based instance-level segmentation for hallucination analysis. In complex scenes, SAM often over-segments the image into many fragmented masks (Fig.~\ref{fig:cfd_sam}), which not only increases the computational cost (Table~\ref{tab:efficiency}) but also makes normal background structures more likely to be misclassified as hallucinated objects.

\paragraph{\textbf{ReMOVE}}
ReMOVE~\cite{chandrasekar2024remove} suffers from two limitations. First, its cropping strategy is not well suited to multi-object removal. As shown in Fig.~\ref{fig:crop_vs}, ReMOVE uses a single enlarged crop even when targets are spatially far apart, which introduces excessive irrelevant background and dilutes the feature difference between removed regions and their surrounding context. In addition, the crop is not guaranteed to be square; resizing it to a fixed square feature resolution may distort object appearance and further affect feature comparison.

Second, ReMOVE can produce counter-intuitive rankings under simple local perturbations. In Fig.~\ref{fig:blur_swap}, we apply Gaussian blur to the masked regions or directly swap the two masked regions. Human judgment clearly prefers the original image, followed by the blurred one, and then the swapped one. However, both ReMOVE and CFD often produce the reverse ordering, whereas RC-S remains consistent with human perception. This behavior may be related to ReMOVE's cropping strategy, and is also consistent with its averaging-based design, which can dilute localized structural corruption.

To further quantify this behavior, we extend the analysis to RORD-Val~\cite{zhao2026objectclear} by comparing each original image against blurred and region-swapped variants. As shown in Table~\ref{tab:rord_perturb}, RC-S consistently preserves the correct ranking in all valid cases, while ReMOVE and CFD often fail, especially under Gaussian blur.

\begin{figure}[!htbp]
    \centering
    \begin{subfigure}{0.8\linewidth}
        \centering
        \includegraphics[width=\linewidth]{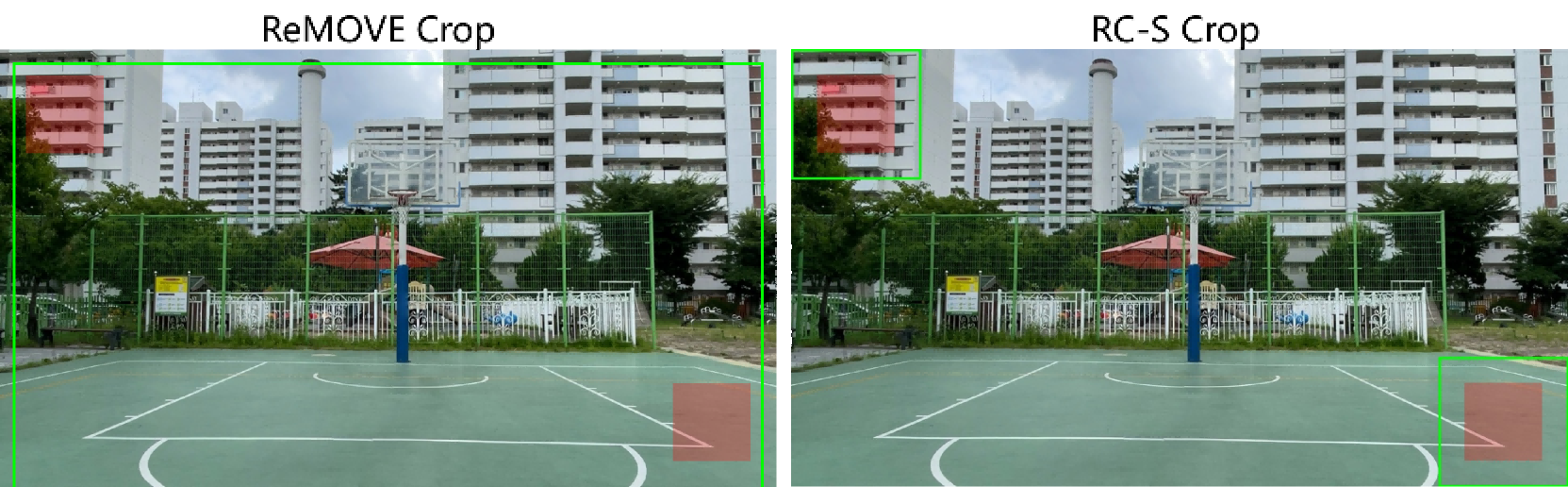}
        \caption{Cropping strategies of ReMOVE and RC-S. \textcolor{green}{Green} boxes denote crop regions, and \textcolor{red}{red} areas denote masks.}
        \label{fig:crop_vs}
    \end{subfigure}

    \begin{subfigure}{0.8\linewidth}
        \centering
        \includegraphics[width=\linewidth]{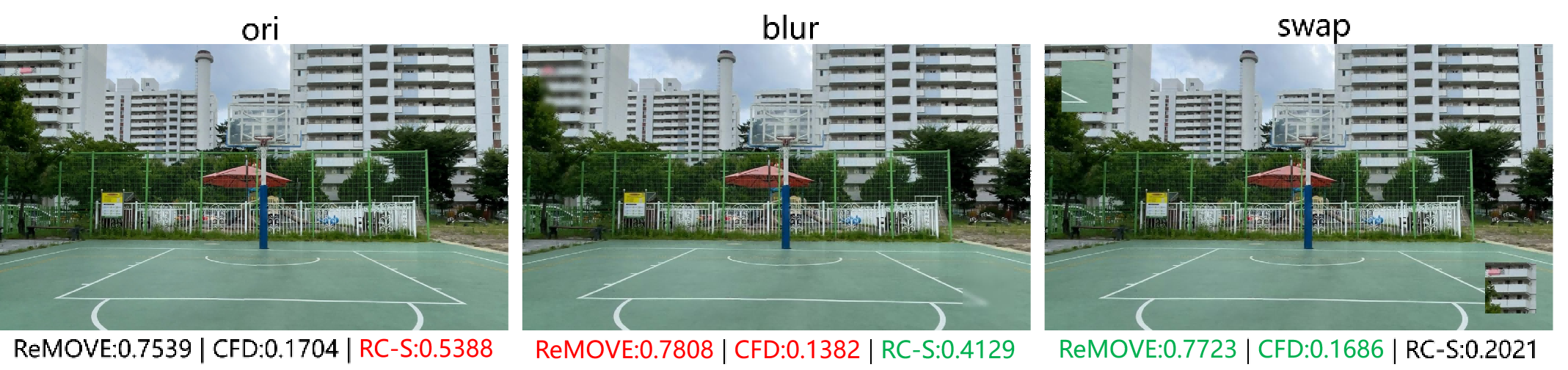}
        \caption{Counter-intuitive rankings under local perturbations. \textcolor{red}{Red} marks the best-ranked result, and \textcolor{green}{green} marks the second-best.}
        \label{fig:blur_swap}
    \end{subfigure}

    \caption{Further analysis of ReMOVE and RC-S. ReMOVE uses a single enlarged crop and produces counter-intuitive rankings, while RC-S remains more consistent with human judgment.}
    \label{fig:remove_flaw}
\end{figure}

\begin{table}[t]
\centering
\caption{Correct preference rates under controlled perturbations on RORD-Val. We report the percentage of cases in which each metric correctly prefers the original image over its perturbed variant. The last column reports the percentage of cases where both ReMOVE and CFD fail.}
\label{tab:rord_perturb}
\begin{tabular}{lcccc}
\toprule
\textbf{Perturbation} & \textbf{RC-S} & \textbf{ReMOVE} & \textbf{CFD} & \textbf{Both Fail} \\
\midrule
Gaussian Blur & 100.00\% & 60.06\% & 49.27\% & 21.28\% \\
Region Swap   & 100.00\% & 75.22\% & 76.97\% & 4.66\% \\
\bottomrule
\end{tabular}
\end{table}

\paragraph{\textbf{TokSim}}
We do not provide case-level visual comparisons for TokSim~\cite{kushwaha2026object}, since it is not publicly released and cannot be reproduced in our pipeline\footnote{Update after submission: TokSim has been open-sourced at \url{https://github.com/sakshamsingh1/object_wiper}.}. Still, its formulation suggests several possible limitations. TokSim relies on cosine similarity and DINOv3 features, which, as discussed in Sec.~\ref{sec:ablation} and Sec.~\ref{sec:supp_fourier_analysis}, may not be the most suitable choices for capturing localized restoration artifacts. Moreover, TokSim multiplies three factors---temporal consistency, dissimilarity to the input object, and similarity to neighboring background---into a single score. Such direct fusion may be overly restrictive, since these dimensions can interact and may not be reliably combined through simple multiplication. As a result, errors in one term may propagate to the final score and reduce correlation with human judgment.

\subsection{Sensitivity to Incomplete Removal}
Incomplete removal leaves residual foreground content inside the edited region and is a critical failure mode for object removal. To evaluate metric sensitivity to such residues, we create controlled variants on RORD-Val by pasting the original foreground back into clean removal results at different mask-area ratios. Table~\ref{tab:residue_preference} reports the percentage of cases in which each metric correctly ranks the clean result above its residue-corrupted counterpart. RC-S remains highly sensitive even at a 30\% residue ratio, whereas ReMOVE and CFD make substantially more incorrect preferences.

\begin{table}
\centering
\caption{Correct preference rates for clean results over residue-corrupted variants on RORD-Val.}
\label{tab:residue_preference}
\begin{tabular}{lccc}
\toprule
\textbf{Residue Ratio} & \textbf{RC-S} & \textbf{ReMOVE} & \textbf{CFD} \\
\midrule
30\% & \textbf{96.5\%} & 73.5\% & 84.8\% \\
50\% & \textbf{99.7\%} & 89.2\% & 93.0\% \\
\bottomrule
\end{tabular}
\end{table}
\section{Comprehensive Details of PROVE-Bench}
\label{sec:supple_bench_details}
\subsection{Dataset Statistics and Construction Overview}
\label{sec:dataset_statis}

We provide more comprehensive statistics and construction details of PROVE-Bench in this section. In particular, besides the benchmark statistics, we include a visual overview of the PROVE-M construction pipeline to complement Sec.~\ref{PROVE-M} in the main paper.

\textbf{PROVE-M} contains 80 motion-augmented paired samples derived from real-world paired recordings. As shown in Table~\ref{tab:dataset_statistics}, its source paired recordings are balanced across several axes, including object count (40 single-object / 40 multi-object), illumination condition (40 bright / 40 low-light or night), target type (60 person / 20 object), and target motion (67 dynamic / 13 static). In addition, PROVE-M also covers challenging real-world factors such as small targets, large-area shadows, and complex side effects, including 52 reflection-related cases.

\newcolumntype{A}{>{\centering\arraybackslash}m{1.45cm}} % ObjCnt / General
\newcolumntype{B}{>{\centering\arraybackslash}m{1.45cm}} % Illum / Dyn.-Bg.
\newcolumntype{C}{>{\centering\arraybackslash}m{1.75cm}} % Type / Textured Bg.
\newcolumntype{D}{>{\centering\arraybackslash}m{1.85cm}} % TgtMotion / Complex Refl.
\newcolumntype{E}{>{\centering\arraybackslash}m{1.20cm}} % Small / Crowd
\newcolumntype{F}{>{\centering\arraybackslash}m{1.95cm}} % Refl.-related / Fast Motion

\begin{table*}
    \centering
    \setlength{\tabcolsep}{4pt}
    \renewcommand{\arraystretch}{1.12}
    \caption{Statistics of \textbf{PROVE-Bench}. \textbf{GT}: target-free ground truth. \textbf{Fmt.}: format. \textbf{L/P}: landscape / portrait.}
    \label{tab:dataset_statistics}
    \resizebox{\linewidth}{!}{
    \begin{tabular}{l | c c c c c | c}
        \toprule
        \textbf{Subset} & \textbf{\#Videos} & \textbf{GT} & \textbf{Motion} & \textbf{Res.} & \textbf{Fmt.} & \textbf{Statistics} \\
        \midrule
        \textbf{PROVE-M}
        & 80
        & \cmark
        & Dynamic
        & 1080p
        & L/P
        & \begin{tabular}{A|B|C|D|E|F}
            Obj. Num. & Illumination & TargetType & Motion & Small & Refl.-related \\
            \hline
            40/40 & 40/40 & 60/20 & 67/13 & 6 & 52
          \end{tabular} \\
        \midrule
        \textbf{PROVE-H}
        & 100
        & \xmark
        & Dynamic
        & 1080p
        & L/P
        & \begin{tabular}{A|B|C|D|E|F}
            General & Dyn.-Bg. & Textured Bg. & Complex Refl. & Crowd & Fast Motion \\
            \hline
            35 & 15 & 20 & 14 & 7 & 9
          \end{tabular} \\
        \bottomrule
    \end{tabular}
    }
\end{table*}

\textbf{PROVE-H} contains 100 real-world videos without target-free ground truth. To characterize its diversity, we group the videos by their primary challenge into general scenes (35), highly textured background scenes (20), dynamic-background scenes (15), complex-reflection scenes (14), fast-motion scenes (9), and crowd scenes (7). Together, these categories cover a broad range of realistic difficulties, including cluttered backgrounds, strong scene dynamics, severe occlusions, and complex physical interactions.

Figure~\ref{fig:prove_m_pipeline} further illustrates the construction pipeline of PROVE-M. Starting from real-world paired recordings, we generate input--mask--ground-truth triplets using SAM~3, perform post-processing including BG-PSNR ranking, mask-difference filtering, and human selection, and finally apply Ken Burns-style motion simulation to obtain motion-augmented paired samples.

\begin{figure}
  \centering
  \includegraphics[width=0.8\linewidth]{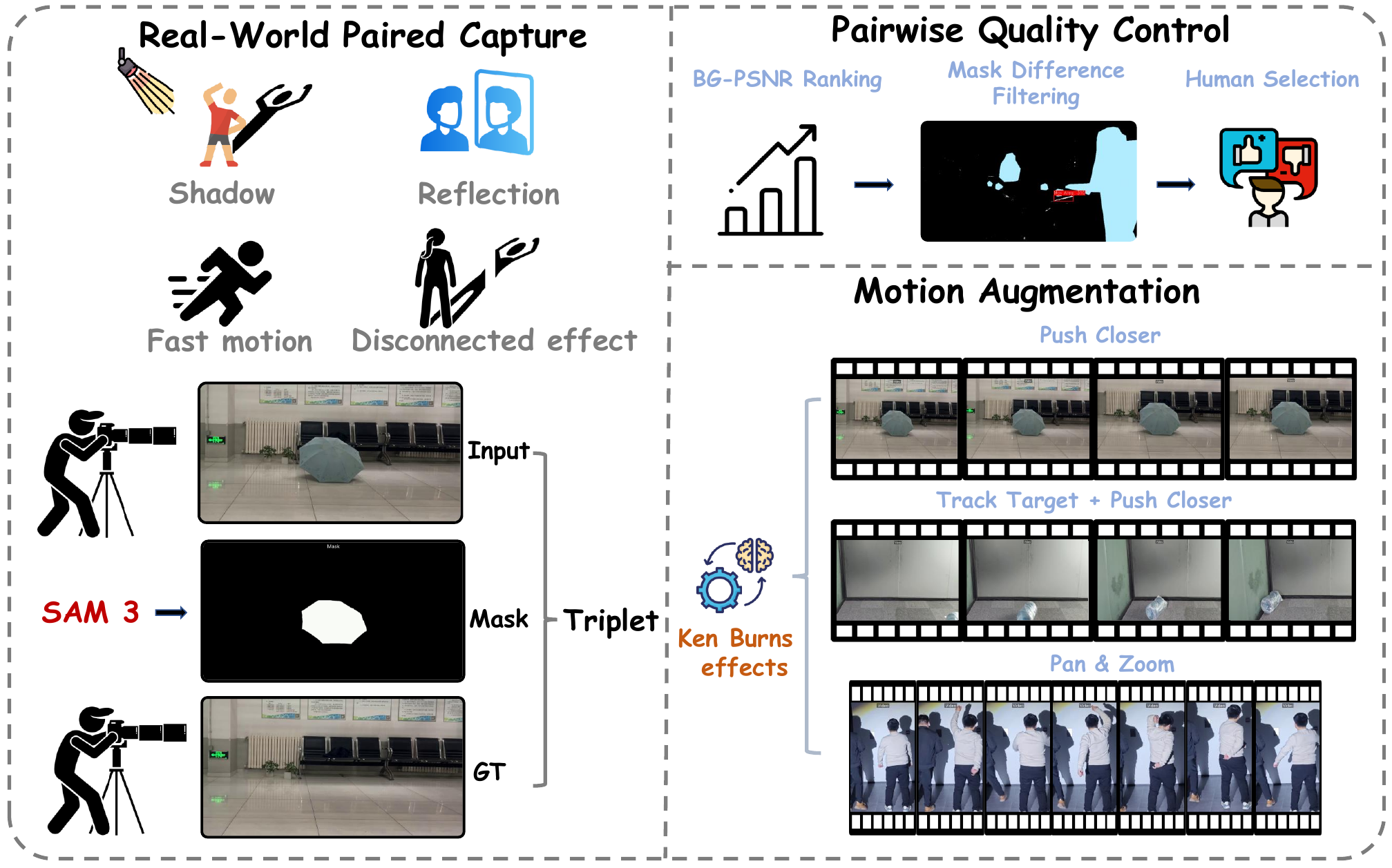}
  \caption{Construction pipeline of PROVE-M.}
  \label{fig:prove_m_pipeline}
\end{figure}

\begin{table*}[t]
    \centering
    \setlength{\tabcolsep}{4.5pt}
    \renewcommand{\arraystretch}{1.08}
    \small

    % non-overlapping annotations
    \newcommand{\drop}[1]{\hspace{0.15em}{\scriptsize\textcolor{red}{$\downarrow$#1}}}
    \newcommand{\badup}[1]{\hspace{0.15em}{\scriptsize\textcolor{red}{$\uparrow$#1}}}
    \newcommand{\goodup}[1]{\hspace{0.15em}{\scriptsize\textcolor{green!60!black}{$\uparrow$#1}}}

    \caption{Comprehensive benchmark results on PROVE-S, PROVE-M, and PROVE-H.
    The values annotated next to the PROVE-M rows indicate the performance change relative to PROVE-S.
    \textcolor{red}{Red} denotes performance degradation under motion augmentation, while \textcolor{green!60!black}{green} denotes slight improvement.
    \textbf{Note:} Since PROVE-H lacks target-free ground truth, its full-reference metrics (PSNR, SSIM, LPIPS) are computed exclusively on the unmasked background regions relative to the original input frames.}
    \label{tab:prove_full_results}

    \resizebox{\textwidth}{!}{
    \begin{tabular}{c l c c c c c c c}
        \toprule
        \textbf{Method} & \textbf{Dataset} & \textbf{PSNR}$\uparrow$ & \textbf{SSIM}$\uparrow$ & \textbf{LPIPS}$\downarrow$ & \textbf{ReMOVE}$\uparrow$ & \textbf{CFD}$\downarrow$ & \textbf{RC-S}$\uparrow$ & \textbf{RC-T}$\downarrow$ \\
        \midrule

        \multirow{3}{*}{Minimax~\cite{zi2025minimax}}
        & PROVE-S & 23.73 & 0.879 & 0.114 & 0.890 & 0.308 & 0.495 & 0.264 \\
        & PROVE-M & 22.06\drop{1.67} & 0.867\drop{0.012} & 0.145\badup{0.031} & 0.872\drop{0.018} & 0.328\badup{0.020} & 0.481\drop{0.014} & 0.445\badup{0.181} \\
        & PROVE-H & 29.94 & 0.890 & 0.059 & 0.858 & 0.372 & 0.463 & 0.322 \\
        \midrule

        \multirow{3}{*}{GenOmni~\cite{lee2025generative}}
        & PROVE-S & 26.89 & 0.883 & 0.108 & 0.900 & 0.337 & 0.529 & 0.056 \\
        & PROVE-M & 25.45\drop{1.44} & 0.881\drop{0.002} & 0.128\badup{0.020} & 0.883\drop{0.017} & 0.371\badup{0.034} & 0.509\drop{0.020} & 0.320\badup{0.264} \\
        & PROVE-H & 27.76 & 0.856 & 0.107 & 0.867 & 0.408 & 0.507 & 0.223 \\
        \midrule

        \multirow{3}{*}{ROSE~\cite{miao2025rose}}
        & PROVE-S & 27.35 & 0.891 & 0.076 & 0.902 & 0.292 & 0.507 & 0.317 \\
        & PROVE-M & 26.67\drop{0.68} & 0.896\goodup{0.005} & 0.091\badup{0.015} & 0.883\drop{0.019} & 0.328\badup{0.036} & 0.492\drop{0.015} & 0.634\badup{0.317} \\
        & PROVE-H & 27.94 & 0.872 & 0.063 & 0.857 & 0.381 & 0.468 & 0.423 \\
        \midrule

    \multirow{3}{*}{VACE~\cite{jiang2025vace}}
        & PROVE-S & 21.43 & 0.869 & 0.139 & 0.736 & 0.384 & 0.323 & 0.323 \\
        & PROVE-M & 19.39\drop{2.04} & 0.843\drop{0.026} & 0.178\badup{0.039} & 0.727\drop{0.009} & 0.474\badup{0.090} & 0.320\drop{0.003} & 0.330\badup{0.007} \\
        & PROVE-H & 27.08 & 0.903 & 0.069 & 0.807 & 0.333 & 0.417 & 0.332 \\
        \midrule

        \multirow{3}{*}{DiffuEraser~\cite{li2025diffueraser}}
        & PROVE-S & 24.01 & 0.880 & 0.120 & 0.886 & 0.298 & 0.488 & 0.336 \\
        & PROVE-M & 22.52\drop{1.49} & 0.870\drop{0.010} & 0.138\badup{0.018} & 0.870\drop{0.016} & 0.318\badup{0.020} & 0.478\drop{0.010} & 0.485\badup{0.149} \\
        & PROVE-H & 32.71 & 0.948 & 0.038 & 0.848 & 0.356 & 0.435 & 0.350 \\
        \midrule

        \multirow{3}{*}{ProPainter~\cite{zhou2023propainter}}
        & PROVE-S & 24.27 & 0.896 & 0.106 & 0.882 & 0.260 & 0.447 & 0.431 \\
        & PROVE-M & 22.52\drop{1.75} & 0.886\drop{0.010} & 0.132\badup{0.026} & 0.862\drop{0.020} & 0.283\badup{0.023} & 0.420\drop{0.027} & 0.619\badup{0.188} \\
        & PROVE-H & 35.22 & 0.964 & 0.037 & 0.836 & 0.308 & 0.385 & 0.436 \\
        \midrule

        \multirow{3}{*}{FGT~\cite{zhang2022flow}}
        & PROVE-S & 24.36 & 0.879 & 0.112 & 0.868 & 0.316 & 0.403 & 0.503 \\
        & PROVE-M & 22.63\drop{1.73} & 0.851\drop{0.028} & 0.157\badup{0.045} & 0.877\goodup{0.009} & 0.346\badup{0.030} & 0.379\drop{0.024} & 0.759\badup{0.256} \\
        & PROVE-H & 34.07 & 0.970 & 0.020 & 0.847 & 0.370 & 0.360 & 0.563 \\
        \bottomrule
    \end{tabular}
    }

    \medskip
    \noindent\parbox{\textwidth}{\raggedright\footnotesize
    \textbf{Note:} Due to compliance requirements, the released PROVE-Bench differs slightly from the one used here, resulting in minor metric discrepancies; however, the overall trends remain consistent. Up-to-date results of existing methods on the released dataset are available on: \url{https://xiaomi-research.github.io/prove/\#results}
    }
\end{table*}

\subsection{Pairwise Quality Control of PROVE-M}
\label{sec:pairwise_control}

This section provides further details on the pairwise quality controls processing for our PROVE-M construction, as mentioned in the main text in Sec.~\ref{PROVE-M}.

\textbf{Stage 1: Mask Consistency Computation.} In this stage, we evaluate the overall structural similarity between the difference-based coarse mask ($M_{\text{diff}}$) and the refined ground-truth mask ($M_{\text{gt}}$). For a video pair with $N$ frames, the video-level consistency score $S$ is calculated as the average Peak Signal-to-Noise Ratio (PSNR) across all frames:
\begin{equation}
    S = \frac{1}{N} \sum_{t=1}^{N} 10 \log_{10} \left( \frac{\text{MAX}_I^2}{\text{MSE}(M_{\text{diff}}^{(t)}, M_{\text{gt}}^{(t)})} \right)
\end{equation}
where $\text{MAX}_I = 255$ is the maximum pixel intensity, and $\text{MSE}(\cdot)$ denotes the Mean Squared Error between the two masks. We rank all candidate video pairs in descending order based on $S$ and retain only the top 40\% for the subsequent stage.

\textbf{Stage 2: Background Disturbance Detection.} Even with high overall PSNR scores, some videos may contain severe localized background artifacts (e.g., unintended moving objects captured in $M_{\text{diff}}$ that are absent in $M_{\text{gt}}$). To identify and filter out these samples, we isolate the artifact regions for each frame $t$ via a saturated subtraction:
\begin{equation}
    M_{\text{artifact}}^{(t)} = \max \left( M_{\text{diff}}^{(t)} - M_{\text{gt}}^{(t)}, 0 \right)
\end{equation}
This operation effectively retains only the regions incorrectly identified as foreground in $M_{\text{diff}}$. Subsequently, we perform connected component analysis (CCA) on $M_{\text{artifact}}^{(t)}$ to measure the maximum contiguous area of these isolated artifacts. If the maximum connected area in any frame of a video exceeds a predefined threshold (e.g., 1000), the entire video pair is considered severely disturbed and is discarded.

\textbf{Stage 3: Human Selection.}
The remaining videos are reviewed by experienced human annotators who have simultaneous access to the input video, the target-free ground-truth video, and both mask videos ($M_{\text{diff}}$ and $M_{\text{gt}}$). Annotators perform frame-by-frame inspection and discard videos exhibiting visible misalignment, color discrepancy, or unintended background changes, yielding a final set of 80 high-quality paired samples.

\subsection{Comprehensive Benchmark Results}
\label{sec:bench_result}
To thoroughly evaluate state-of-the-art video object removal models and validate the difficulty of PROVE-Bench, we conduct extensive experiments across three settings: PROVE-S (the original static paired recordings before motion augmentation), PROVE-M (motion-augmented paired), and PROVE-H (hard unconstrained). Quantitative results are summarized in Table~\ref{tab:prove_full_results}.

A direct comparison between the static PROVE-S and the motion-augmented PROVE-M strongly demonstrates the critical impact of camera motion on removal quality. As annotated by the red indicators in Table ~\ref{tab:prove_full_results}, introducing simulated camera movements (e.g., zooming, panning, and shaking) leads to a consistent and significant performance degradation across all evaluated models. This performance gap highlights the necessity of PROVE-M in bridging the evaluability gap between static laboratory captures and authentic, shaky user videos.

Across the benchmarks, diffusion-based methods (e.g., ROSE~\cite{miao2025rose}, GenOmni~\cite{lee2025generative}) generally demonstrate superior spatial contextual coherence compared to traditional methods (e.g., FGT~\cite{zhang2022flow}), as reflected by their higher RC-S scores. Notably, in the highly challenging PROVE-H dataset, while models maintain relatively high background-only PSNR/SSIM scores (indicating that unmasked regions are well-preserved), their RC-S and RC-T scores reveal underlying spatial hallucinations and temporal flickering within the complex erased regions. This further corroborates that existing models have not yet fully resolved the perception-distortion tradeoff in unconstrained environments.

Beyond the quantitative results, the representative cases from PROVE-M and PROVE-H shown in Fig.~\ref{fig:prove_case_m} and Fig.~\ref{fig:prove_case_h} further reveal the limitations of current object removal models in realistic scenarios. The PROVE-M case shows that current models remain weak and unstable in removing large-area shadows associated with multiple foreground targets. The PROVE-H case further shows that highly textured scenes are still very challenging, often resulting in severe residuals, artifacts, and structural distortion after removal.

\begin{figure}[t]
    \centering
    \begin{subfigure}{0.55\linewidth}
        \centering
        \includegraphics[width=\linewidth]{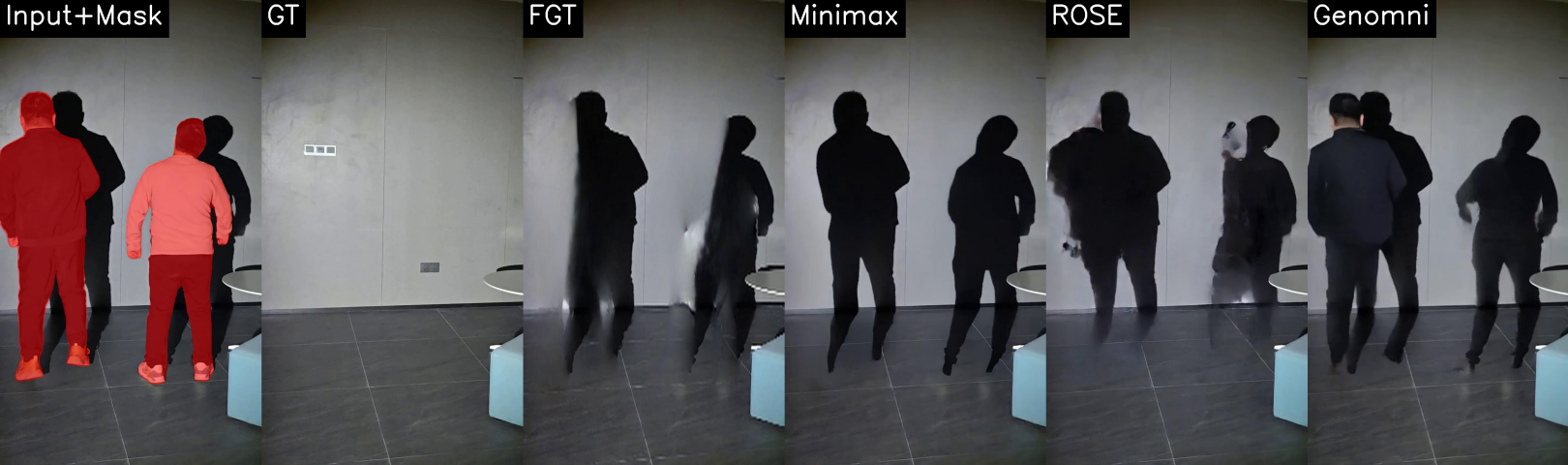}
        \caption{PROVE-M}
        \label{fig:prove_case_m}
    \end{subfigure}
    \hfill
    \begin{subfigure}{0.435\linewidth}
        \centering
        \includegraphics[width=\linewidth]{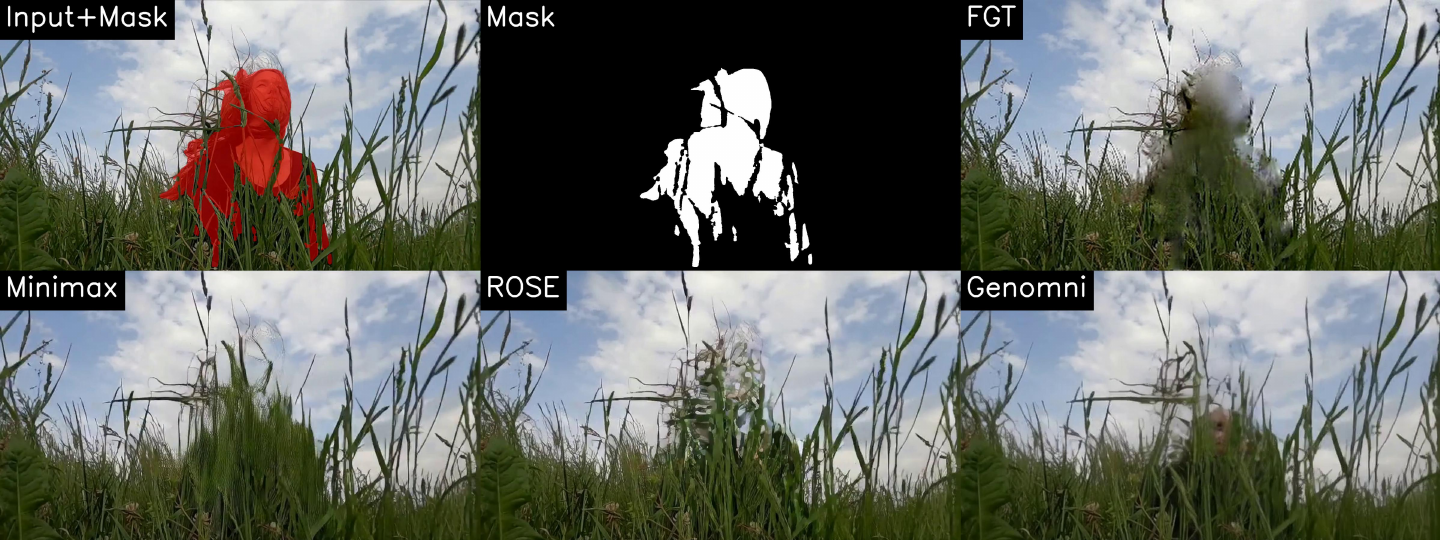}
        \caption{PROVE-H}
        \label{fig:prove_case_h}
    \end{subfigure}

    \caption{Visual results of existing SOTA models for representative cases in PROVE-M and PROVE-H.}
    \label{fig:prove_case}
\end{figure}

\section{Robustness of Human and Automated Evaluation}
\label{sec:evaluation details}
\subsection{Inter-Rater Reliability}

Fig.~\ref{fig:overall_image} shows the visual interface used in our human evaluation, where outputs of all compared models are displayed side by side for direct comparison and ranking.

\begin{figure}[t]
    \centering
    \begin{subfigure}{0.49\linewidth}
        \centering
        \includegraphics[width=\linewidth]{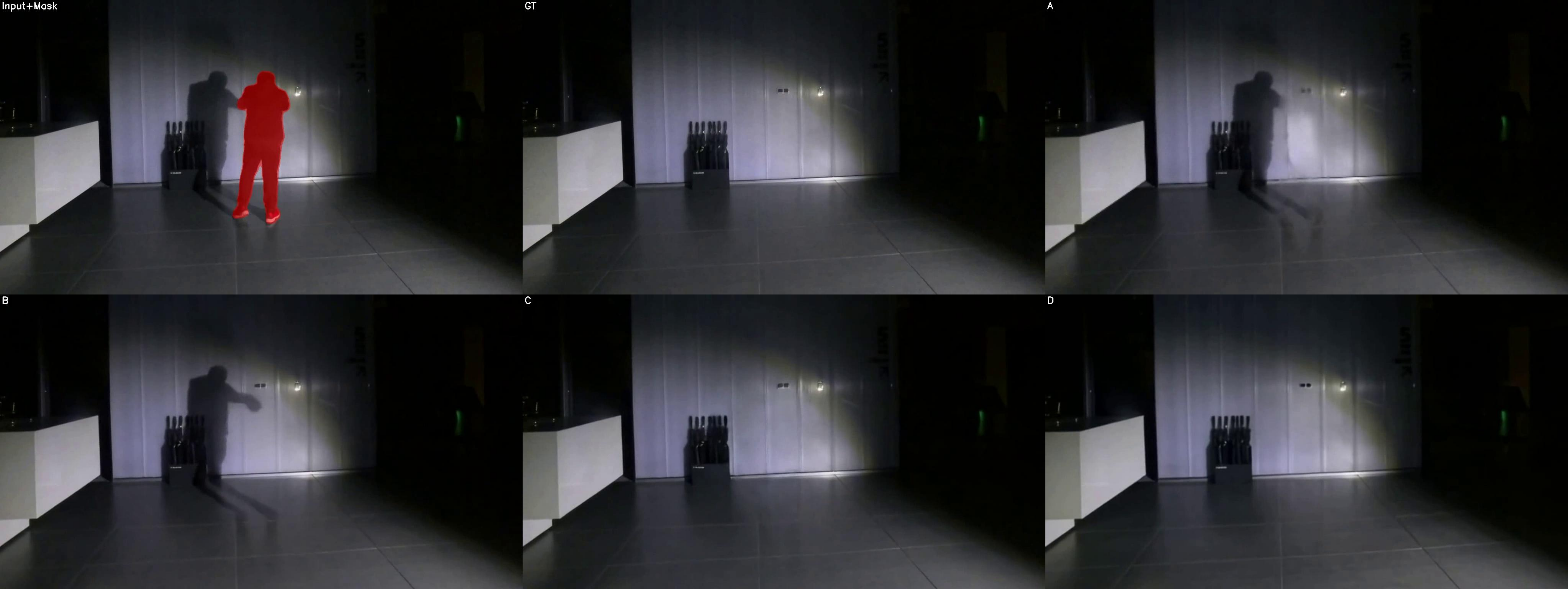}
        \caption{PROVE-M}
        \label{fig:sub_a}
    \end{subfigure}
    \hfill
    \begin{subfigure}{0.49\linewidth}
        \centering
        \includegraphics[width=\linewidth]{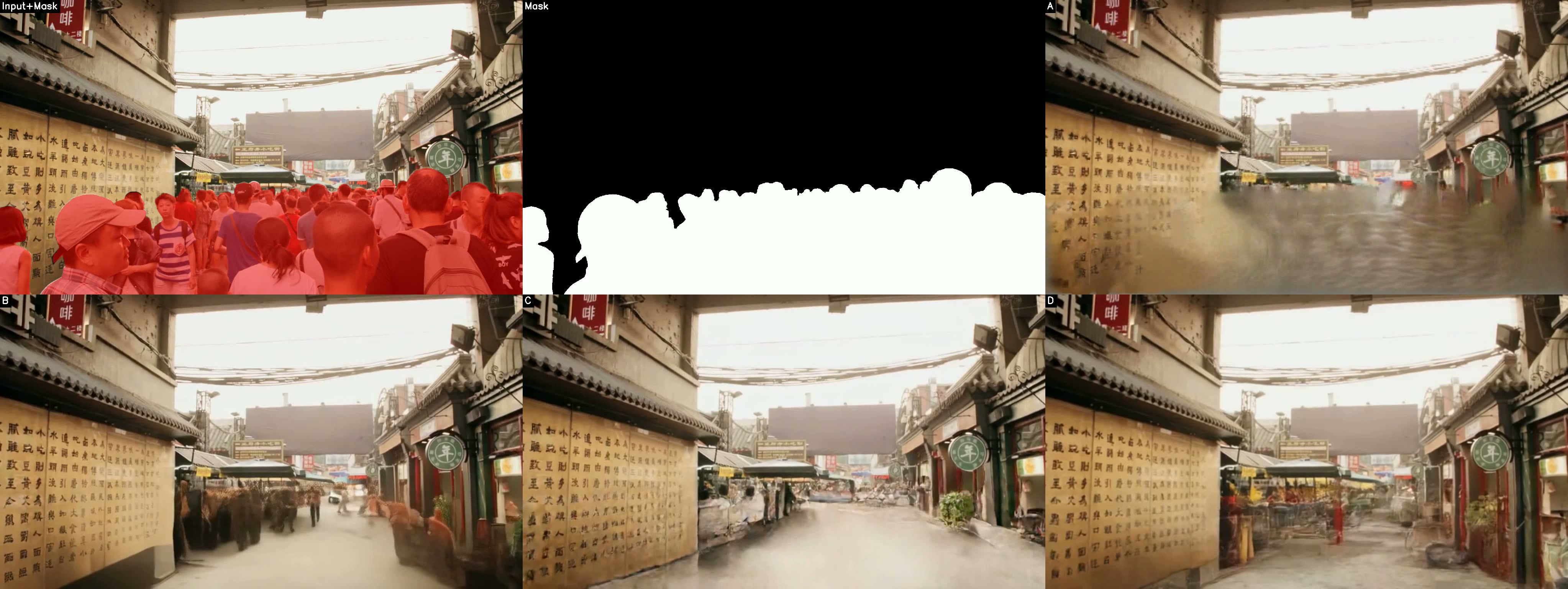}
        \caption{PROVE-H}
        \label{fig:sub_b}
    \end{subfigure}
    \caption{The visual interface for human evaluation. It illustrates the side-by-side comparison of different models on the PROVE-M (a) and PROVE-H (b) datasets, presenting the exact perspective viewed by human evaluators.}
    \label{fig:overall_image}
\end{figure}

To quantify inter-rater agreement, we compute Kendall's coefficient of concordance ($W$) across all annotators. As shown in Table~\ref{tab:concordance}, the agreement scores range from 0.57 to 0.76 across different benchmarks. Notably, the lowest agreement appears on RORD, which aligns with the relatively lower correlation of RC-S on this dataset. This is because RORD contains relatively simple removal cases where even human annotators struggle to distinguish subtle quality differences among models, resulting in lower scoring consistency and consequently lower metric correlation.

\begin{table}[t]
    \centering
    \caption{Kendall's coefficient of concordance ($W$) of human scores across benchmarks.}
    \label{tab:concordance}
    % \setlength{\tabcolsep}{4pt}
    % \footnotesize
    \begin{tabular}{lcccccc}
        \toprule
        & \textbf{RORD} & \textbf{OBER-Wild} & \textbf{DAVIS} & \textbf{ROSE} & \textbf{PROVE-M} & \textbf{PROVE-H} \\
        \midrule
        $W$ & 0.5691 & 0.7574 & 0.6721 & 0.7444 & 0.6666 & 0.6627 \\
        \bottomrule
    \end{tabular}
\end{table}

\subsection{Statistical Reliability of Metric--Human Correlations}
To quantify the uncertainty of the correlations reported in the main paper, we perform 10,000 bootstrap resamples of the human-ranking data for PROVE-M and PROVE-H. Table~\ref{tab:bootstrap_correlation} shows that the confidence intervals for RC-S remain well above zero for both Kendall's $\tau$ and Spearman's $\rho$, with all tests significant at $p<0.001$. These results support that RC-S's strong alignment with human judgments on the two proposed benchmarks is robust to sampling variation.

\begin{table}[htbp]
    \centering
    \caption{Bootstrap reliability of RC-S correlations with human rankings (10,000 resamples).}
    \label{tab:bootstrap_correlation}
    \begin{tabular}{lcccccc}
        \toprule
        Dataset & Kendall's $\tau$ & 95\% CI & Sig. & Spearman $\rho$ & 95\% CI & Sig. \\
        \midrule
        PROVE-M & 0.699 & [0.621, 0.772] & $p<0.001$ & 0.755 & [0.679, 0.825] & $p<0.001$ \\
        PROVE-H & 0.763 & [0.706, 0.817] & $p<0.001$ & 0.823 & [0.775, 0.868] & $p<0.001$ \\
        \bottomrule
    \end{tabular}
\end{table}

\subsection{Challenge-wise Robustness on PROVE-H}
\label{sec:prove_h_challenge_eval}

Beyond aggregate statistical reliability, we examine whether RC-S remains aligned with human judgments across the diverse challenges covered by PROVE-H. We partition PROVE-H into six scenario types---textured backgrounds, crowds, fast motion, general scenes, dynamic backgrounds, and complex reflections---and compute Kendall's $\tau$ between RC-S-induced rankings and aggregated human preferences within each type.

As shown in Table~\ref{tab:prove_h_challenge_corr}, RC-S achieves consistently strong correlations across all challenge types, ranging from 0.722 to 0.827. Although dynamic backgrounds and complex reflections are particularly difficult because their local appearance changes rapidly or contains intertwined side effects, RC-S retains high agreement with human judgments in both settings. These results indicate that the overall PROVE-H correlation is not driven by a single relatively easy scenario; rather, RC-S remains effective across diverse real-world removal challenges.

\begin{table}[htbp]
    \centering
\caption{Challenge-wise Kendall's $\tau$ between RC-S and aggregated human preferences on PROVE-H.}    \label{tab:prove_h_challenge_corr}
    \begin{tabular}{lcccccc}
        \toprule
        Scenario & Tex. Bg. & Crowd & Fast Mot. & General & Dyn. Bg. & Comp. Refl. \\
        \midrule
        Kendall's $\tau$ & 0.827 & 0.785 & 0.758 & 0.750 & 0.739 & 0.722 \\
        \bottomrule
    \end{tabular}
\end{table}

\subsection{Representative Model-Pair Analysis}
To complement the analysis in the main paper, we provide the full Kendall's Tau results for representative model pairs across all benchmarks in Table~\ref{tab:model_pair}. Specifically, we compare GenOmni vs.\ FGT on video benchmarks and ObjectClear vs.\ LaMa on image benchmarks. The results are consistent with our main findings: RC-S achieves the highest and most stable correlation with human judgments, while traditional FR metrics and their variants exhibit inconsistent or negative correlations.

\begin{table}
    \centering
    \caption{Kendall's Tau between each metric and human evaluation on representative model pairs (GenOmni vs.\ FGT for video benchmarks, ObjectClear vs.\ LaMa for image benchmarks).}
    \label{tab:model_pair}
    \begin{tabular}{l cccccc}
        \toprule[1.2pt]
        \textbf{Metric} & \textbf{RORD} & \textbf{OBER-Wild} & \textbf{DAVIS} & \textbf{ROSE} & \textbf{PROVE-M} & \textbf{PROVE-H} \\
        \midrule

        PSNR         &  -0.003 & --        & --        &  0.100  &  0.351  & --        \\
        SSIM         &  -0.773 & --        & --        & -0.133 &  0.455  & --        \\
        LPIPS        &  -0.534 & --        & --        & -0.167 &  0.247  & --        \\
        \midrule

        \rowcolor{gray!8} m-PSNR   &  -0.254 & --        & --        &  0.633  &  0.536  & --        \\
        \rowcolor{gray!8} m-SSIM   &  -0.006 & --        & --        &  \underline{0.800} &  {0.766} & --        \\
        \rowcolor{gray!8} m-LPIPS  &  {0.236} & --        & --        &  \textbf{0.900} &  \underline{0.844} & --        \\
        \midrule

        bg-PSNR      &  -0.271 & -0.993 & -0.889 & -0.200 &  0.247  & -0.939 \\
        bg-SSIM      &  -0.854 & -0.993 & -0.889 & -0.367 &  0.195  & -0.980 \\
        bg-LPIPS     &  -0.854 & -0.993 & -0.889 & -0.667 & -0.117 & -0.939 \\
        \midrule

        \rowcolor{gray!8} ReMOVE   &  \underline{0.417} &  \underline{0.762} &  0.005 &  0.300 &  0.273 &  \underline{0.408} \\
        \rowcolor{gray!8} CFD      &  $-$0.038 &  0.589 &  \underline{0.289} &  0.033 &  0.079 &  0.052 \\
        \midrule

        \rowcolor{cyan!6} RC-S     &  \textbf{0.639} &  \textbf{0.927} &  \textbf{0.867} &  0.767 &  \textbf{0.920} &  \textbf{0.959} \\
        \bottomrule[1.2pt]
    \end{tabular}
\end{table}

Furthermore, to validate the reliability of our human evaluations, we report the Kendall's coefficient of concordance ($W$) among human raters across all benchmarks in Table~\ref{tab:pair_concordance}. The consistently high concordance values (ranging from 0.7901 to 0.9264) demonstrate a strong inter-rater agreement, confirming that the human judgments used as the gold standard in our experiments are highly robust and reliable.

\begin{table}
    \centering
    \caption{Kendall's coefficient of concordance ($W$) of human scores for the representative model-pair analysis across all benchmarks.}
    \label{tab:pair_concordance}
    \begin{tabular}{lcccccc}
        \toprule
        & \textbf{RORD} & \textbf{OBER-Wild} & \textbf{DAVIS} & \textbf{ROSE} & \textbf{PROVE-M} & \textbf{PROVE-H} \\
        \midrule
        $W$ & 0.8722 & 0.9010 & 0.8437 & 0.7901 & 0.9264 & 0.9216 \\
        \bottomrule
    \end{tabular}
\end{table}

\subsection{Exploratory Comparison with GPT-Based Evaluation}
\label{sec:eval_gpt}
In addition to human evaluation, we explore using GPT-4o as a complementary automated evaluator to assess model rankings. The detailed prompt used for GPT-based evaluation is provided in Fig.~\ref{fig:gpt_prompts}. For each image pair, GPT-4o is asked to rate the removal result from three perspectives: target removal accuracy, visual naturalness, and physical \& detail integrity, each on a scale of 1 to 5. The final GPT score is obtained by averaging the three sub-scores. Compared to human evaluation, GPT-based evaluation offers better scalability and reproducibility while avoiding potential annotator fatigue and subjective bias. As shown in Table~\ref{tab:gpt_test}, RC-S achieves the best or second-best alignment with GPT rankings on most benchmarks, demonstrating that its assessment is consistent with both human and large multimodal model-based evaluation.

\begin{table*}[!htbp]
    \centering
    \caption{Correlation with GPT rankings. We report Kendall's $\tau$ and Average Spearman correlation $\rho$ between metric-induced rankings and aggregated GPT rankings. Best and second-best results are in \textbf{bold} and \underline{underlined}, respectively. ``--'' denotes unavailable metrics.}
    \label{tab:gpt_test}
    \resizebox{\textwidth}{!}{
        \renewcommand{\arraystretch}{1.15}
        \begin{tabular}{l cccccc @{\hspace{2em}} cccccc}
            \toprule[1.2pt]
            \multirow{2.5}{*}{\textbf{Metric}} & \multicolumn{6}{c}{\textbf{Kendall's Tau}} & \multicolumn{6}{c}{\textbf{Average Spearman Correlation}} \\
            \cmidrule(lr){2-7} \cmidrule(l){8-13}
            & RORD & OBER-Wild & DAVIS & ROSE & PROVE-M & PROVE-H & RORD & OBER-Wild & DAVIS & ROSE & PROVE-M & PROVE-H \\
            \midrule

            PSNR         &  \underline{0.2157} & --      & --      &  0.2323 &  0.2680 & --
                         &  \underline{0.2468} & --      & --      &  0.2959 &  0.3039 & --      \\
            SSIM         & -0.0250 & --      & --      &  0.0368 &  0.4278 & --
                         & -0.0434 & --      & --      &  0.0570 &  0.4723 & --      \\
            LPIPS        &  0.0127 & --      & --      &  0.1590 &  0.3730 & --
                         &  0.0096 & --      & --      &  0.1970 &  0.4320 & --      \\
            \midrule
            \rowcolor{gray!8} m-PSNR       & -0.2228 & --      & --      &  0.3167 &  0.3391 & --
                              & -0.2529 & --      & --      &  0.3935 &  0.4034 & --      \\
            \rowcolor{gray!8} m-SSIM       & -0.0075 & --      & --      &  0.4172 & \underline{0.4982} & --
                              & -0.0140 & --      & --      &  0.4953 &  {0.5782} & --      \\
            \rowcolor{gray!8} m-LPIPS      &  0.0849 & --      & --      & \textbf{0.5399} & \underline{0.4982} & --
                              &  0.1065 & --      & --      & \textbf{0.6058} & \underline{0.5884} & --      \\
            \midrule

            bg-PSNR      &  0.0819 & -0.3319 & -0.4094 &  0.1101 &  0.1768 & -0.3732
                         &  0.0916 & -0.4012 & -0.4680 &  0.1526 &  0.2280 & -0.4424 \\
            bg-SSIM      & -0.0508 & -0.1265 & -0.4131 & -0.0410 &  0.2986 & -0.4010
                         & -0.0718 & -0.1631 & -0.4657 & -0.0497 &  0.3223 & -0.4691 \\
            bg-LPIPS     & -0.0888 & -0.2550 & -0.4724 & -0.1243 &  0.1313 & -0.3842
                         & -0.1083 & -0.3183 & -0.5280 & -0.1264 &  0.1595 & -0.4623 \\
            \midrule

            \rowcolor{gray!8} ReMOVE  &  0.1284 &  \underline{0.4457} &  {0.1391} &  0.1656 &  0.1978 &  \underline{0.1782}
                              &  0.1492 &  \underline{0.5006} &  \underline{0.1863} &  0.1959 &  0.2130 &  \underline{0.2162} \\
            \rowcolor{gray!8} CFD   &  0.0403 &  0.3834 &  \underline{0.1569} &  0.0029 &  0.2294 &  0.1240
                              &  0.0466 &  0.4430 &  0.1863 &  0.0184 &  0.2629 &  0.1238 \\
            \midrule

            \rowcolor{cyan!6} RC-S  & \textbf{0.2280} &  \textbf{0.4689} & \textbf{0.5461} & \underline{0.5183} & \textbf{0.5015} & \textbf{0.4469}
                              & \textbf{0.2627} &  \textbf{0.5467} & \textbf{0.6276} & \underline{0.5677} & \textbf{0.5991} & \textbf{0.5202} \\
            \bottomrule[1.2pt]
        \end{tabular}
    }
\end{table*}

\ifarxiv
    \begin{figure*}[htbp]
    \begin{tcolorbox}[
        title=Prompt for GPT-based Evaluation,
        colback=gray!8,
        colframe=black!75,
        fonttitle=\bfseries,
        arc=2mm,
        fontupper=\small,
        boxsep=1mm,
        left=3mm, right=3mm, top=2mm, bottom=2mm,
        toptitle=1mm, bottomtitle=1mm
    ]
    \textbf{Human:} You are a professional image rater evaluating object removal edits. You will be given two images:

    \begin{itemize}
        \setlength{\itemsep}{2pt}
        \item The first image is the original photo before editing. The red translucent area indicates the region to be removed, including all foreground objects that need to be eliminated.
        \item The second image is the result after object removal editing.
    \end{itemize}

    The objects removed may be humans, physical objects, text, or a combination of these.  \\
    Your task is to evaluate the object removal quality from \textbf{three perspectives}, each on a scale of 1 to 5.  \\
    For each score, provide a clear and specific explanation. Then, identify the removed object category and any new objects that were mistakenly generated.

    \vspace{0.8em}
    \hrule
    \vspace{0.8em}

    \textbf{Target Removal Accuracy (1--5):} Did the system remove the correct object(s) completely, and only the intended object(s)?
    \begin{itemize}
        \setlength{\itemsep}{0pt}
        \item[\textbf{1 --}] Nothing removed, or unrelated object removed.
        \item[\textbf{2 --}] Target only partly removed, or wrong object/class removed, or new unintended object appears.
        \item[\textbf{3 --}] Target mostly removed, but fragments remain, or extra objects were also removed.
        \item[\textbf{4 --}] Only intended object(s) removed, but with minor collateral loss (e.g., nearby detail lost, count incorrect).
        \item[\textbf{5 --}] Perfect: all and only the specified object(s) removed; everything else preserved precisely.
    \end{itemize}

    \textbf{Visual Naturalness (1--5):} How natural and seamless does the edited image look?
    \begin{itemize}
        \setlength{\itemsep}{0pt}
        \item[\textbf{1 --}] Severely broken (holes, artifacts, glitches, disfigured areas).
        \item[\textbf{2 --}] Obvious erase marks, mismatched texture/color, or blurry/aliased patches.
        \item[\textbf{3 --}] Acceptable, but has visible inconsistencies in lighting, texture, or resolution.
        \item[\textbf{4 --}] Mostly natural with only minor artifacts visible when closely inspected.
        \item[\textbf{5 --}] Seamless and photorealistic; indistinguishable from an unedited image.
    \end{itemize}

    \textbf{Physical \& Detail Integrity (1--5):} Is the structure, geometry, and realism of the scene preserved?
    \begin{itemize}
        \setlength{\itemsep}{0pt}
        \item[\textbf{1 --}] Physically implausible result (e.g., floating limbs, warped objects, broken perspective).
        \item[\textbf{2 --}] Major geometry or detail disruption; background structure lost or repeated unnaturally.
        \item[\textbf{3 --}] Mostly consistent perspective and lighting; minor issues localized.
        \item[\textbf{4 --}] Geometry and scene preserved; background logically filled in.
        \item[\textbf{5 --}] Completely realistic and coherent, with well-restored structure, lighting, and fine details.
    \end{itemize}

    \textbf{Additional Tasks:}
    \begin{itemize}
        \setlength{\itemsep}{0pt}
        \item Identify the category/class of the object that was removed by comparing the two images.
        \item If any unreasonable or unintended new objects are generated in the result, list their class(es); if none, write \texttt{none}.
    \end{itemize}

    \textbf{Important:} The second and third scores must \textbf{NOT} exceed the first score.

    \vspace{0.8em}
    \hrule
    \vspace{0.8em}

    \textbf{\#\#\# Final Output Format (strictly follow this):}

    \begin{ttfamily}
    Target Removal Accuracy: <score 1--5>  \\
    Visual Naturalness: <score 1--5> \\
    Physical \& Detail Integrity: <score 1--5>  \\
    Erase object class: <e.g., human, text, bottle, backpack, person+text> \\
    Generate object class: <e.g., blur patch, ghost figure, floating object>, or `none' \\[1ex]
    <Image> Image 1 </Image> \\
    <Image> Image 2 </Image>
    \end{ttfamily}

    \vspace{0.5em}
    \textbf{GPT:} ...
    \end{tcolorbox}
    \vspace{-2mm}
    \caption{Prompt for GPT-based evaluation.}
    \label{fig:gpt_prompts}
    \end{figure*}

\else

    \begin{figure*}[t]
    \centering
    \setlength{\fboxsep}{5pt}
    \fbox{%
    \begin{minipage}{0.96\textwidth}
    \small
    \textbf{Prompt for GPT-based Evaluation}

    \vspace{0.4em}
    \hrule
    \vspace{0.6em}

    \textbf{Human:} You are a professional image rater evaluating object removal edits. You will be given two images:

    \begin{itemize}
        \setlength{\itemsep}{2pt}
        \item The first image is the original photo before editing. The red translucent area indicates the region to be removed, including all foreground objects that need to be eliminated.
        \item The second image is the result after object removal editing.
    \end{itemize}

    The objects removed may be humans, physical objects, text, or a combination of these.  \\
    Your task is to evaluate the object removal quality from \textbf{three perspectives}, each on a scale of 1 to 5.  \\
    For each score, provide a clear and specific explanation. Then, identify the removed object category and any new objects that were mistakenly generated.

    \vspace{0.8em}
    \hrule
    \vspace{0.8em}

    \textbf{Target Removal Accuracy (1--5):} Did the system remove the correct object(s) completely, and only the intended object(s)?
    \begin{itemize}
        \setlength{\itemsep}{0pt}
        \item[\textbf{1 --}] Nothing removed, or unrelated object removed.
        \item[\textbf{2 --}] Target only partly removed, or wrong object/class removed, or new unintended object appears.
        \item[\textbf{3 --}] Target mostly removed, but fragments remain, or extra objects were also removed.
        \item[\textbf{4 --}] Only intended object(s) removed, but with minor collateral loss (e.g., nearby detail lost, count incorrect).
        \item[\textbf{5 --}] Perfect: all and only the specified object(s) removed; everything else preserved precisely.
    \end{itemize}

    \textbf{Visual Naturalness (1--5):} How natural and seamless does the edited image look?
    \begin{itemize}
        \setlength{\itemsep}{0pt}
        \item[\textbf{1 --}] Severely broken (holes, artifacts, glitches, disfigured areas).
        \item[\textbf{2 --}] Obvious erase marks, mismatched texture/color, or blurry/aliased patches.
        \item[\textbf{3 --}] Acceptable, but has visible inconsistencies in lighting, texture, or resolution.
        \item[\textbf{4 --}] Mostly natural with only minor artifacts visible when closely inspected.
        \item[\textbf{5 --}] Seamless and photorealistic; indistinguishable from an unedited image.
    \end{itemize}

    \textbf{Physical \& Detail Integrity (1--5):} Is the structure, geometry, and realism of the scene preserved?
    \begin{itemize}
        \setlength{\itemsep}{0pt}
        \item[\textbf{1 --}] Physically implausible result (e.g., floating limbs, warped objects, broken perspective).
        \item[\textbf{2 --}] Major geometry or detail disruption; background structure lost or repeated unnaturally.
        \item[\textbf{3 --}] Mostly consistent perspective and lighting; minor issues localized.
        \item[\textbf{4 --}] Geometry and scene preserved; background logically filled in.
        \item[\textbf{5 --}] Completely realistic and coherent, with well-restored structure, lighting, and fine details.
    \end{itemize}

    \textbf{Additional Tasks:}
    \begin{itemize}
        \setlength{\itemsep}{0pt}
        \item Identify the category/class of the object that was removed by comparing the two images.
        \item If any unreasonable or unintended new objects are generated in the result, list their class(es); if none, write \texttt{none}.
    \end{itemize}

    \textbf{Important:} The second and third scores must \textbf{NOT} exceed the first score.

    \vspace{0.8em}
    \hrule
    \vspace{0.8em}

    \textbf{\#\#\# Final Output Format (strictly follow this):}

    \begin{ttfamily}
    Target Removal Accuracy: <score 1--5>  \\
    Visual Naturalness: <score 1--5> \\
    Physical \& Detail Integrity: <score 1--5>  \\
    Erase object class: <e.g., human, text, bottle, backpack, person+text> \\
    Generate object class: <e.g., blur patch, ghost figure, floating object>, or `none' \\[1ex]
    <Image> Image 1 </Image> \\
    <Image> Image 2 </Image>
    \end{ttfamily}

    \vspace{0.6em}
    \hrule
    \vspace{0.6em}

    \textbf{GPT:} ...
    \end{minipage}%
    }
    \caption{Prompt for GPT-based evaluation.}
    \label{fig:gpt_prompts}
    \end{figure*}

\fi
\section{Extended Experimental Results}
\label{sec:Extended Experimental Results}

\begin{table}[htbp]
    \centering
    \caption{Computational cost comparison.}
    \label{tab:efficiency}
    \setlength{\tabcolsep}{10pt}
    \begin{tabular}{llr}
        \toprule
        \textbf{Type} & \textbf{Metric} & \textbf{Time} \\
        \midrule
        \multirow{3}{*}{\textit{Spatial Metrics}}
            & ReMOVE & 180.7 ms/frame \\
            & CFD    & 1842.8 ms/frame \\
            & \cellcolor{cyan!6}\textbf{RC-S (Ours)} & \cellcolor{cyan!6}{134.6 ms/frame} \\
        \midrule
        \multirow{3}{*}{\textit{Temporal Metrics}}
            & TC     & 47.9 ms/frame-pair \\
            & TF     & 22.3 ms/frame-pair \\
            & \cellcolor{cyan!6}\textbf{RC-T (Ours)} & \cellcolor{cyan!6}{183.8 ms/frame-pair} \\
        \bottomrule
    \end{tabular}
\end{table}

\subsection{Runtime and Computational Cost}
Table~\ref{tab:efficiency} reports the computational cost of all metrics on a single NVIDIA 4090 GPU. Spatial metrics are measured on $448\times448$ images, and temporal metrics are measured on 81-frame videos at $448\times448$ resolution. Among spatial metrics, RC-S is the most efficient at 134.6\,ms per frame, 13.7$\times$ faster than CFD. For temporal metrics, all metrics operate on pairs of adjacent frames and are measured in ms/frame-pair. RC-T is slower than TC and TF, as its sliding-window mechanism focuses on finer local details for more thorough temporal quality assessment.

\subsection{Extended Results for Figure 2}
We supplement Figure 2 with extended results of the {ROSE}~\cite{miao2025rose} method under varying diffusion inference steps on the DAVIS dataset, as shown in Fig.~\ref{fig:fig2_suppl}. Consistent with the observations in the main paper, FR metrics such as PSNR and SSIM increase as the number of inference steps decreases, despite a clear degradation in visual quality.

\begin{figure}[t]
    \centering
    \includegraphics[width=\linewidth]{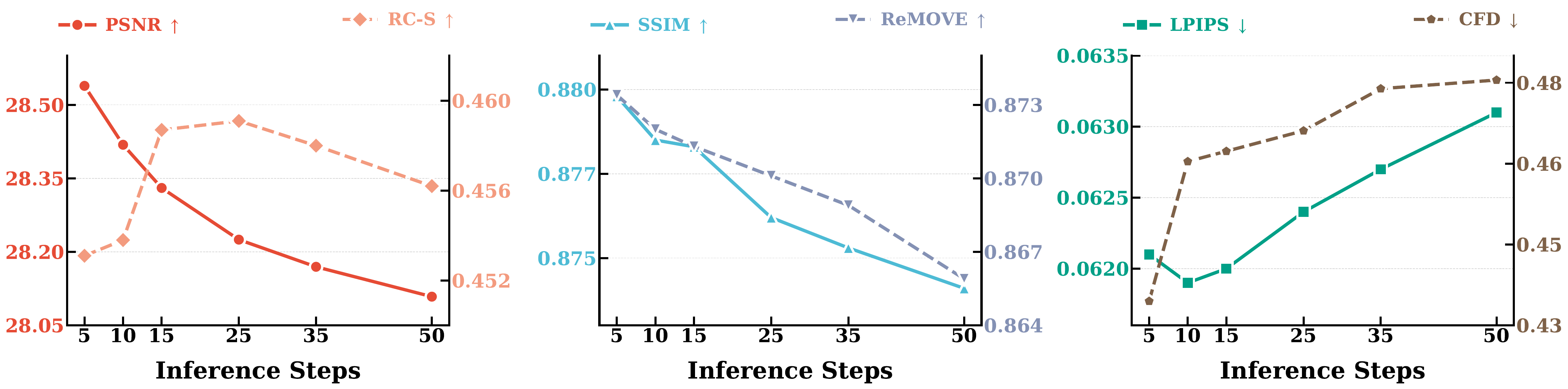}
    \caption{Metric responses to different inference steps for ROSE~\cite{miao2025rose} on DAVIS.}
    \label{fig:fig2_suppl}
\end{figure}

\begin{figure}
    \centering
    \includegraphics[width=\linewidth]{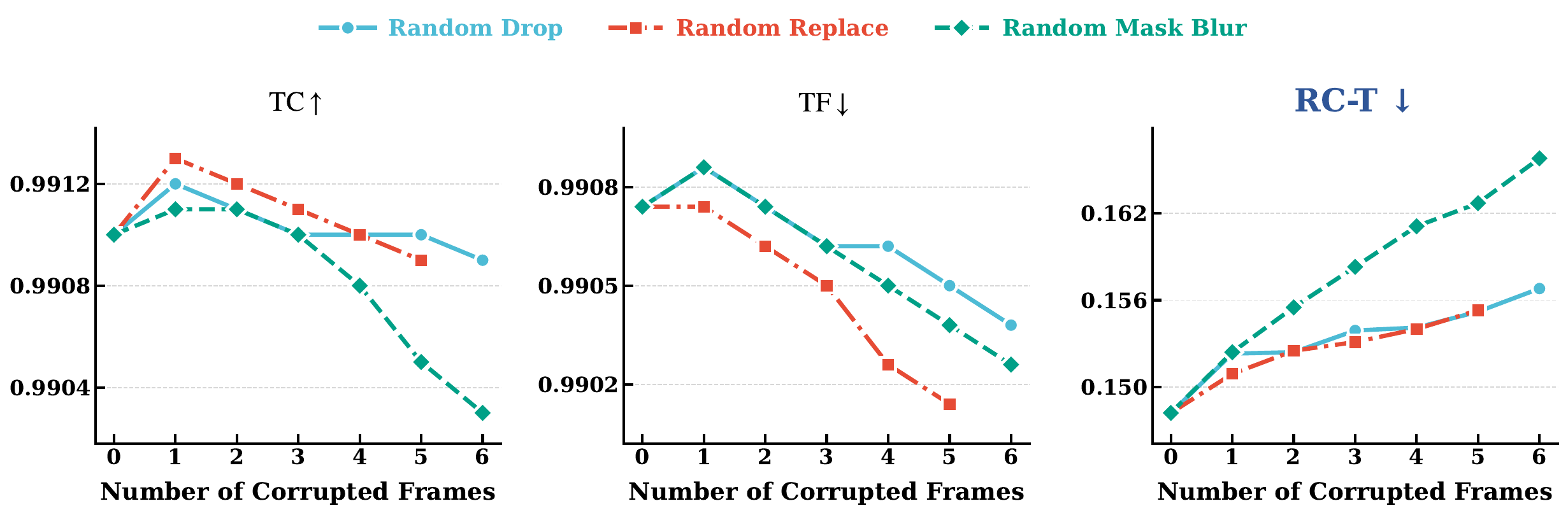}
    \caption{Sensitivity of temporal metrics to Random Drop, Random Replace, and Random Mask Blur corruptions on ROSE-Bench.}
    \label{fig:rct_suppl}
\end{figure}

\subsection{Extended Results for Fig.~\ref{fig:both_datasets}}
\label{sec:extended_fig5}
We extend the temporal robustness analysis of Fig.~\ref{fig:both_datasets} to the ROSE-Bench dataset. We perform \textbf{Random Drop}, \textbf{Random Replace}, and \textbf{Random Mask Blur} corruptions and progressively increase the number of corrupted frames. Specifically, \textbf{Random Mask Blur} applies a Gaussian blur exclusively within the target mask region of the selected frames to simulate localized temporal degradation. To avoid incidental results, the corrupted frame indices are fixed for each video across different corruption levels, while remaining different across videos. The corrupted frames are also accumulated progressively, so that higher corruption levels include all corrupted frames from lower levels. As shown in Fig.~\ref{fig:rct_suppl}, the results are consistent with Fig.~\ref{fig:both_datasets} in the main paper: TC and TF remain insensitive to all three corruption types, while RC-T exhibits a clear monotonic response to increasing corruption levels, further confirming its superior sensitivity to localized temporal artifacts.

\subsection{Ablation Study of RC-T}
\label{RC-T Ablation}
Building upon the ablation study of RC-S, which has thoroughly validated the effectiveness of the core components, we further investigate the RC-T metric. Specifically, we conduct an ablation to verify the necessity of the crop operation. As illustrated in Fig.~\ref{fig:crop_ablation}, without the crop operation, the metric is easily dominated by large background areas. Consequently, it becomes severely insensitive to temporal disruptions, failing to penalize even obvious anomalies such as the randomly replaced frames introduced in our experiments. By incorporating the crop operation, RC-T focuses exclusively on the target area, significantly enhancing its sensitivity to actual temporal incoherence.

\begin{figure}[t]
    \centering
    \includegraphics[width=0.4\linewidth]{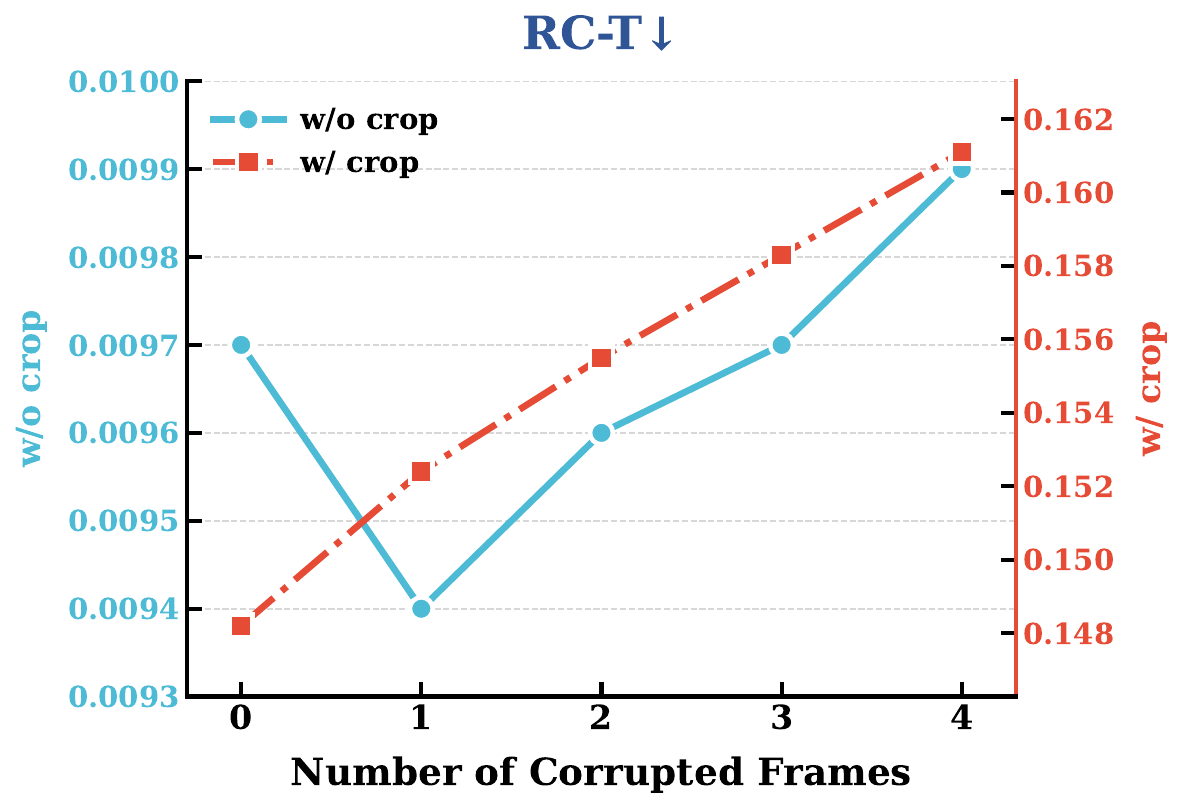}
    \caption{Ablation study on the RC-T.}
    \label{fig:crop_ablation}
\end{figure}

\subsection{Additional Ablation Study of RC-S}
\label{Additional Ablation}
Under the default configuration, we set the MMD kernel size to 10 and the sliding-window size to one quarter of the feature-map resolution. Figure~\ref{fig:rcs_hyperparameter_ablation} jointly evaluates the sensitivity of RC-S to the kernel size, window size, and target bounding-box expansion factor. RC-S remains stable over a practical range of kernel and window settings. It also performs consistently across local expansion ratios, with the best result at one third of the box size, but degrades substantially when evaluated on the full image. These results confirm that local cropping, rather than a precisely tuned expansion ratio, is the essential design choice.

\begin{figure*}[t]
    \centering
    \includegraphics[width=1.0\linewidth]{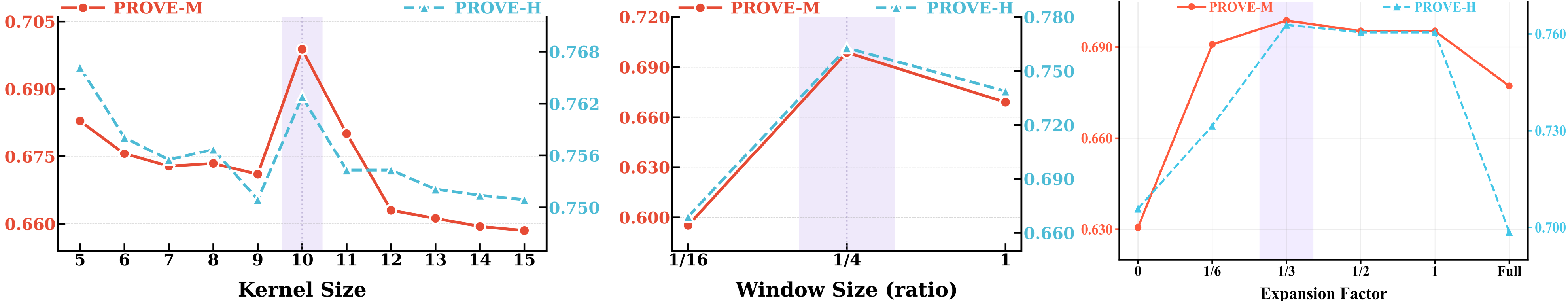}
    \caption{Hyperparameter sensitivity of RC-S with respect to the MMD kernel size, sliding-window size, and target bounding-box expansion factor. RC-S is stable across practical local settings, whereas full-image evaluation degrades performance.}
    \label{fig:rcs_hyperparameter_ablation}
\end{figure*}

\section{Extended Discussion on Limitations}
\label{sec:Extended Discussion on Limitations}

\begin{figure}
  \centering
  \includegraphics[width=\linewidth]{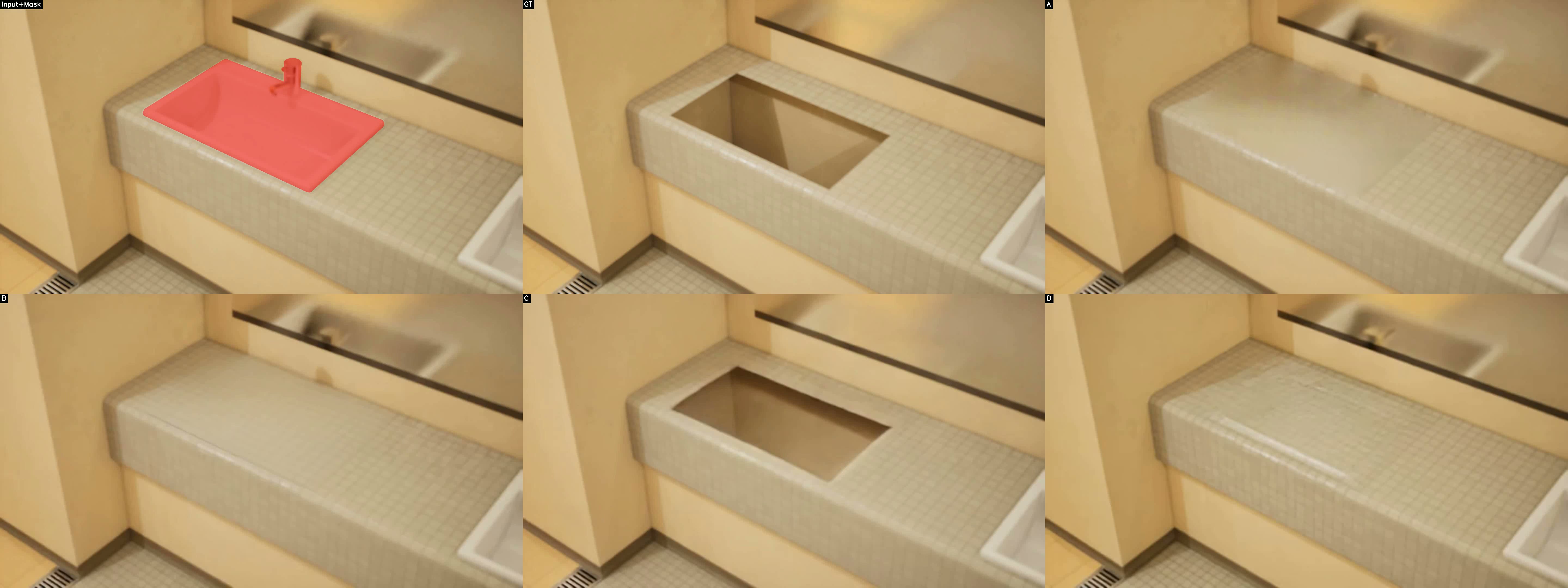}
  \caption{One-to-many ambiguity in object removal. Although the predictions differ from the provided GT, they can still be visually plausible.}
  \label{fig:limitation}
\end{figure}

\begin{figure}
  \centering
  \includegraphics[width=\linewidth]{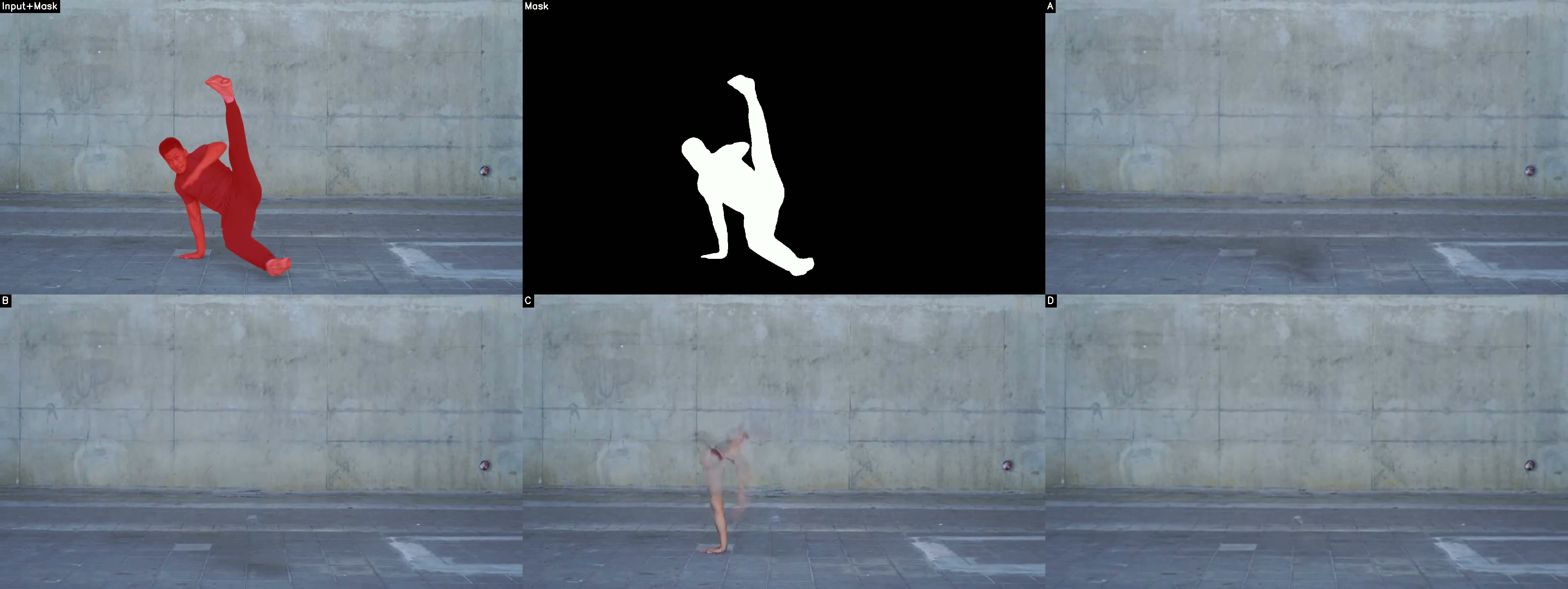}
  \caption{Example showing that RC-S can still capture locally visible side effects. A/B/C/D rankings (1 = best): RC-S = 3/2/4/1, ReMOVE = 1/3/4/2, CFD = 1/2/4/3.}
  \label{fig:limitation_side}
\end{figure}

Our work has three main limitations. First, object removal is inherently a one-to-many problem: multiple restored results can be visually plausible even if they differ from a single target-free reference. Accordingly, our metrics are designed to favor contextual coherence and perceptual plausibility rather than exact agreement with a particular GT. As illustrated in Fig.~\ref{fig:limitation}, for the sink-removal case from ROSE-Bench, the reference preserves the recessed cavity after the sink is removed, while several models instead generate a tiled countertop. Although these results deviate from the provided GT, they can still appear visually reasonable. Therefore, our metric may prefer a contextually coherent result that is not the closest one to a specific reference.

Second, our metric has limited spatial coverage for large side effects. Since the evaluation is performed on cropped local regions, extended side effects such as large shadows or reflections may not always be fully covered. As a result, the metric may underestimate errors whose spatial extent goes beyond the cropped evaluation region. Still, this does not mean that RC-S is insensitive to side effects in general. When side-effect-related degradation is locally visible, RC-S can still capture it effectively. As illustrated in Fig.~\ref{fig:limitation_side}, RC-S correctly ranks the results according to residual content, artifacts, and side effects, in agreement with human perception.

Third, our benchmark still cannot fully reproduce real free-camera recording effects, such as 3D viewpoint changes, motion blur, and rolling shutter.Given the practical difficulty of collecting paired target-present and target-free videos under unconstrained real-world camera motion, we approximate such scenarios through motion simulation. A possible future direction is to use cameras mounted on robotic arms with predefined motion trajectories, which may enable the collection of paired data under more realistic natural camera movement.

\fi

\end{document}